\documentclass{article} 
\usepackage{iclr2025_conference,times}

\usepackage{amsmath,amsfonts,bm}

\def\eqref#1{equation~\ref{#1}}

\def\1{\bm{1}}

\DeclareMathAlphabet{\mathsfit}{\encodingdefault}{\sfdefault}{m}{sl}
\SetMathAlphabet{\mathsfit}{bold}{\encodingdefault}{\sfdefault}{bx}{n}

\usepackage[hidelinks,colorlinks=true,linkcolor=cyan,citecolor=cyan]{hyperref}
\usepackage{url}

\usepackage{multirow}
\usepackage{makecell}
\usepackage{tabularx}
\usepackage{graphicx}
\usepackage{booktabs}
\usepackage{xcolor}
\usepackage{color,colortbl}
\usepackage{wrapfig}
\usepackage{subcaption}

\title{iFormer: Integrating ConvNet and Transformer for Mobile Application}

\author{Chuanyang Zheng \\
Independent Researcher \\
\texttt{chuanyang\_zheng@sjtu.edu.cn}
}

\iclrfinalcopy 
\begin{document}

\maketitle

\begin{abstract}
	We present a new family of mobile hybrid vision networks, called iFormer,
	with a focus on optimizing latency and accuracy on mobile applications. iFormer effectively integrates the fast local representation capacity of convolution with the efficient global modeling ability of self-attention. The local interactions are derived from transforming a standard convolutional network, \textit{i.e.}, ConvNeXt, to design a more lightweight mobile network. Our newly introduced mobile modulation attention removes memory-intensive operations in MHA and employs an efficient modulation mechanism to boost dynamic global representational capacity. We conduct comprehensive experiments demonstrating that iFormer outperforms existing lightweight networks across various tasks.
	Notably, iFormer achieves an impressive Top-1 accuracy of 80.4\% on ImageNet-1k with a latency of only 1.10 ms on an iPhone 13,
	surpassing the recently proposed MobileNetV4 under similar latency constraints. Additionally, our method shows significant improvements in downstream tasks, including COCO object detection, instance segmentation, and ADE20k semantic segmentation, while still maintaining low latency on mobile devices for high-resolution inputs in these scenarios. Code and models are available at: \href{https://github.com/ChuanyangZheng/iFormer}{https://github.com/ChuanyangZheng/iFormer}.
\end{abstract}

\section{Introduction}
Building lightweight neural networks facilitates real-time analysis of images and videos captured by mobile applications such as smartphones. 
This not only enhances privacy protection and security by processing data locally on the device but also improves overall user experience.
\begin{wrapfigure}{r}{6.5cm}
	\centering
	\vspace{-2mm}
	\includegraphics[width=\linewidth]{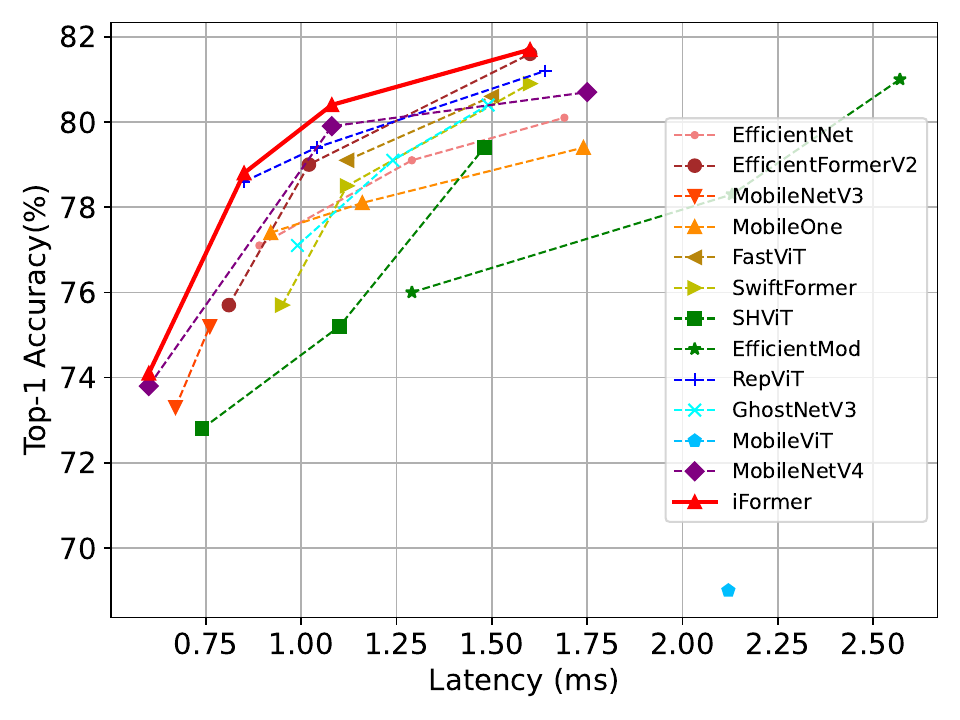}
	\vspace{-6mm}
	\caption{\textbf{Comparison of latency and accuracy between our iFormer and other existing methods on ImageNet-1k.} The latency is measured on an iPhone 13. Our iFormer is Pareto-optimal.}
	\label{fig:comparison}
	\vspace{-2mm}
\end{wrapfigure}
Through the decades, convolutional neural networks (CNNs)~\citep{krizhevsky2012imagenet, szegedy2015going, he2016deep} have emerged as the primary choice for balancing latency and performance on resource-constrained mobile devices. 
However, a significant limitation of CNNs is their reliance on a local sliding window mechanism, which imposes crucial inductive biases that may hinder modeling flexibility.
Recently, the soaring development of vision transformers (ViTs)~\citep{dosovitskiy2020image} has begun to dominate various computer vision tasks, including image classification~\citep{zhai2022scaling}, object detection~\citep{liu2021swin}, and semantic segmentation~\citep{xie2021segformer}. The core mechanism underlying ViTs is self-attention, which dynamically learns interactions between all image patches. This enables the model to focus on important regions adaptively and capture more global features. Nevertheless, deploying ViTs on mobile devices with limited resources poses significant challenges. On the one hand, the quadratic computational complexity of attention renders them unsuitable for large feature maps, which are common in the early stages of vision networks. On the other hand, the multi-head mechanism requires reshaping operations, leading to increased memory usage. 

Many research efforts are devoted to combining the advantages of both CNNs and ViTs in designing lightweight networks while mitigating inefficient operations in mobile applications. Some studies~ \citep{zhang2023rethinking, wang2024repvit, ma2024efficient} revisit the architectural designs of lightweight CNNs from a ViT perspective and incorporate key components that contribute to the performance of ViTs into CNNs. Although these pure lightweight CNNs show improved performance compared to previous mobile networks~\citep{howard2017mobilenets, zhang2018shufflenet, sandler2018mobilenetv2}, they still lag behind the powerful self-attention in ViTs. 
Another line of works~\citep{mehta2021mobilevit, chen2022mobile, li2023rethinking, cai2023efficientvit, shaker2023swiftformer, vasu2023fastvit, qin2024mobilenetv4} 
proposes innovative attention mechanisms to address the limitation of standard attention~\citep{vaswani2017attention} and blend convolutions to achieve a better balance between latency and performance. These attention mechanisms either reduce the number of queries and keys~\citep{shaker2023swiftformer, qin2024mobilenetv4}, limit the attention span~\citep{wan2023seaformer}, or adopt linear attention~\citep{cai2023efficientvit}, which may compromise performance to some extent.

In this work, we present the iFormer, a herd of lightweight models that integrates the strengths of both CNNs and ViTs, achieving a state-of-the-art balance between latency and accuracy. Specifically, we employ a hierarchical architecture consisting of four stages. In the earlier, high-resolution stages, we utilize fast convolution to extract local representations. To construct the convolutional block, we start with a ``modern" ConvNeXt~\citep{liu2022convnet}, which incorporates a series of design decisions inspired by ViTs. Then we progressively ``lighten" the ConvNeXt to create a streamlined lightweight network, optimizing it for real-time mobile latency on an iPhone 13, in contrast to the FLOPs and parameters used in prior works~\citep{mehta2021mobilevit, chen2022mobile}. This results in a fast convolutional architecture with strong performance. To further enhance the dynamic properties and its ability to model long-range contexts, we incorporate self-attention in the later low-resolution stages. However, direct implementation of standard multi-head self-attention (MHA) brings notable memory overheads and slows down inference speed on mobile devices. We identify that the increased latency stems primarily from the reshaping operations in MHA. More analyses reveal that multiple attention heads behave similarly. Therefore, we propose a simple yet effective single-head modulation self-attention (SHMA), which significantly minimizes memory costs while preserving strong performance. Fig.~\ref{fig:arch} provides an illustration of SHMA. In detail, SHMA learns spatial context interactions through optimized self-attention. Concurrently, a parallel feature extraction branch is employed to capture informative features. Finally, we fuse the outputs of these two branches to facilitate a more flexible and dynamic exchange of information, compensating for the slight performance degradation of the single-head attention when compared to MHA.

Benefiting from the fast local representation capacity of convolution and the efficient global modeling proficiency of the proposed SHMA, iFormer outperforms existing pure lightweight CNNs and hybrid networks across multiple visual recognition tasks, including image classification, object detection, instance segmentation, and semantic segmentation. For instance, in the context of image classification as shown in Fig.~\ref{fig:comparison}, iFormer-M achieves a Top-1 accuracy of 80.4\% with only 1.10 ms on an iPhone 13 without advanced training strategies such as knowledge distillation~\citep{touvron2021training} or reparameterization~\citep{ding2021repvgg}. Notably, our model obtains a 0.5\% improvement in Top-1 accuracy compared to the recent MNV4-Conv-M~\citep{qin2024mobilenetv4}, while being 1.4$\times$ faster than FastViT-SA12~\citep{vasu2023fastvit} with similar accuracy. These results demonstrate the effectiveness of the proposed network in capturing both local and global feature representations.

\section{Related Work}
\subsection{Efficient Convolutional Networks}
In the past 2010s, computer vision was dominated by CNNs, and so were efficient networks. The first remarkable breakthrough in mobile CNNs is MobileNets~\citep{howard2017mobilenets}, which hatches the concept of decomposing standard convolution into depthwise and pointwise counterparts. Subsequently, MobileNetV2~\citep{sandler2018mobilenetv2} introduces an inverted residual bottleneck block to push the state-of-the-art for mobile models. Numerous studies have aimed to accelerate CNNs via various approaches, such as channel shuffle in ShuffleNet~\citep{zhang2018shufflenet, ma2018shufflenet} and cheap linear transformations in GhostNet~\citep{han2020ghostnet}. Meanwhile, Neural architecture search (NAS) has emerged as a method for automating the design of neural networks, optimizing for performance under resource constraints. EfficientNet~\citep{tan2019efficientnet}, MobileNetV3~\citep{howard2019searching}, and FBNet~\citep{wu2019fbnet} all achieve rather good performance. Besides, MobileOne~\citep{vasu2023mobileone} proposes to train a model using reparameterizable branches, which are merged during inference. Recently, following the revolution of ViTs, several methods reexamine the design spaces and training strategies~\citep{liu2024ghostnetv3} for mobile CNNs. For instance, RepViT~\citep{wang2024repvit} integrates efficient architectural designs from ViTs into MobileNetV3, outperforming existing lightweight CNNs.
Other approaches, such as FocalNet~\citep{yang2022focal}, Conv2Former~\citep{hou2024conv2former}, and EfficientMod~\citep{ma2024efficient}, fuse features from context modeling and feature projection branches, also known as modulation mechanism, to enhance the model with dynamic properties analogous to attention.
However, pure CNNs remain inherently spatially localized and their reliance on stationary weights restricts their flexibility. Although modulation can partially mitigate this limitation by enhancing dynamic capacity, they still exhibit deficiencies in building global interactions.
\subsection{Efficient Vision Transformers}
The success of Vision Transformer~\citep{dosovitskiy2020image} offers a compelling demonstration of the potential to apply transformer to computer vision tasks. Following this, ViT and its numerous variants~\citep{liu2021swin, dong2022cswin, li2022mvitv2} sweep across various scenarios. However, the quadratic complexity of self-attention behind ViTs poses significant challenges for efficiency. The following researches seek to boost ViT efficiency through efficient attention mechanisms~\citep{wang2021pyramid, zhu2023biformer, hatamizadeh2023fastervit}, model compression~\citep{liu2021post, zheng2022savit}, knowledge distillation~\citep{hao2021learning}, and token reduction~\citep{rao2021dynamicvit, bolya2022token}. Recent studies further introduce ViTs into mobile applications. One mainstream of work combines efficient convolution and ViT to create lightweight hybrid networks~\citep{mehta2022separable, vasu2023fastvit}. MobileViT~\citep{mehta2021mobilevit} directly integrates MobileNetv2 blocks and ViT blocks, while Mobile-Former~\citep{chen2022mobile} features a parallel design of MobileNet and ViT with a two-way bridge connecting the two.
To further accelerate inference, some approaches replace the standard attention~\citep{vaswani2017attention} with efficient variants within the hybrid networks. These include reducing the number of delegate tokens for computing attention~\citep{pan2022edgevits}, employing channel attention~\citep{maaz2022edgenext}, substituting projection in attention with efficient ghost modules~\citep{ma2022mocovit}, and utilizing linear attention mechanisms~\citep{zhao2022lightweight}.
Besides manual designs, EfficientFormer~\citep{li2022efficientformer, li2023rethinking} and MobileNetV4~\citep{qin2024mobilenetv4} search for efficient architectures in a unified space encompassing both convolution operators and transformer operators. Another stream of work focuses on efficient attention mechanisms and directly employs them throughout the entire network~\citep{shaker2023swiftformer, cai2023efficientvit}. For example, CMT~\citep{guo2022cmt} takes advantage of depth-wise convolution to downsample key and value to reduce computation.
GhostNetV2~\citep{tang2022ghostnetv2} applies two fully connected layers along the horizontal and vertical directions to compute attention, a decoupled version of MLP-Mixer~\citep{tolstikhin2021mlp}. Recently, SHViT observes computational redundancy in the multi-head attention module and proposes to apply sing-head attention. In contrast to these existing approaches, we introduce a novel efficient attention module without sacrificing informative interactions, thereby maintaining strong representational capacity. Regarding attention design, ours is a bit similar to SHViT but is considerably superior as shown in Table~\ref{tab:app image classification} in the supplementary material. The key difference lies in the novel modulation attention. In addition, we explore efficient attention mechanisms in an on-device environment while SHViT focuses on general-purpose GPUs, fundamentally different hardware.
\section{Method} \label{sec:method}
\begin{figure}[t]
	\centering
	\begin{minipage}{0.57\textwidth}
			\includegraphics[width=\linewidth]{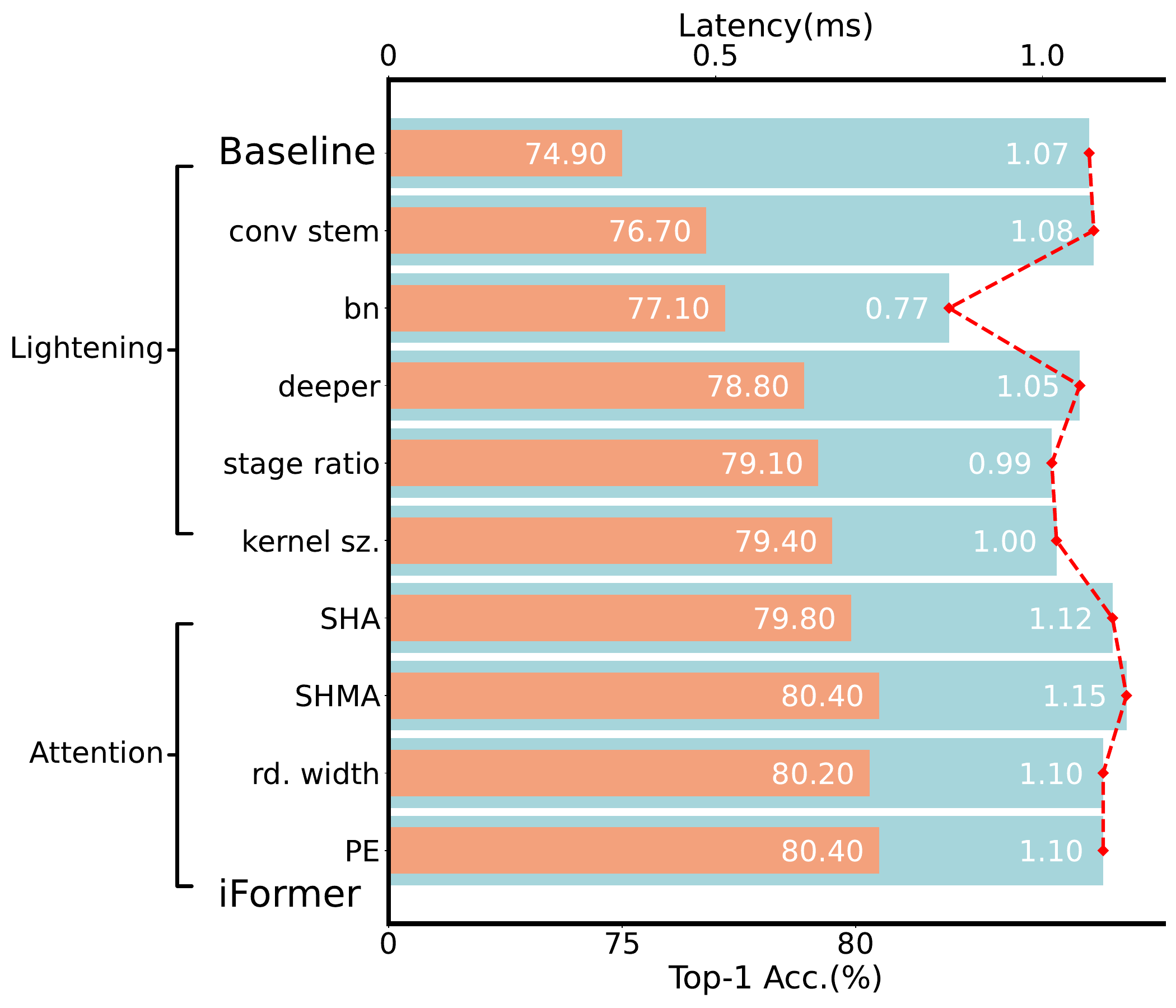}
			\vspace{-7mm}
			\caption{\textbf{Illustration of the evolution from the ConvNeXt baseline towards the lightweight iFormer.} The orange bars are model accuracies and the light blue bars are model latencies. We also include a red latency outline for better visualization.}
			\label{fig:roadmap}
			\vspace{-2mm}
		\end{minipage}
	\hspace{0.01\textwidth}
	\begin{minipage}{0.40\textwidth}
			\centering
			\includegraphics[width=\linewidth]{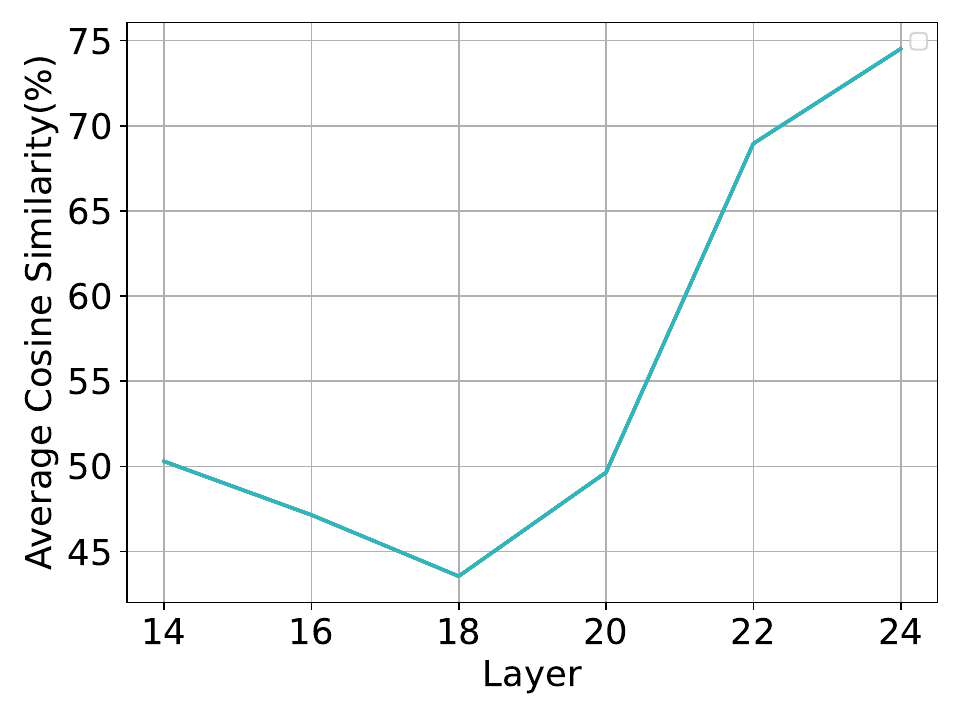}
			\vspace{-7mm}
			\caption{\textbf{The distribution of average cosine similarity among multiple heads within the MHA mechanism.} As the layer depth increases, the similarity goes higher.}
			\label{fig:similarity}
			\vspace{1.5mm}
			\begin{minipage}{0.97\textwidth}
					\centering
					\captionof{table}{\textbf{Latency comparison between multi-head and single-head baseline.}}
					\small
					\vspace{-2mm}
					\setlength{\tabcolsep}{2.5pt}
					\scalebox{0.85}{
							\begin{tabular}{ccc}
									\toprule
									Models & Latency (ms) & Top-1 Acc. (\%)\\
									\midrule
									MHA Baseline & 1.40 & 79.9 \\
									SHA Baseline & 1.12 (1.25$\times$) & 79.8\\
									\bottomrule
								\end{tabular}
						}
				\label{tab:single vs. multi head}
				\end{minipage}
			\vspace{-2mm}
		\end{minipage}
	\vspace{-2mm}
\end{figure}

We present the overall architecture of our iFormer in Fig.~\ref{fig:arch}, which offers a Pareto-optimal accuracy-latency trade-off on mobile applications.
Our exploration towards a streamlined lightweight network unfolds as follows: 1) establishing the baseline and measure metric in Sec.~\ref{sec:prepare}.
2) exploring acceleration techniques consisting of macro and micro designs in Sec.~\ref{sec:macro_micro}.
3) injecting global attention in Sec.~\ref{sec:shma}. Finally, we create a new family of efficient hybrid vision transformers tailored for mobile applications in Sec.~\ref{sec:iformer}. 
A detailed trajectory illustrating the evolution from a general hierarchical CNN to a fast hybrid vision transformer is depicted in Fig.~\ref{fig:roadmap}.

\subsection{Preparing ConvNeXt} \label{sec:prepare}
Our goal is to create an efficient multiscale network, where spatial dimensions of intermediate representations shrink as inference proceeds. In this hierarchical architecture, early network layers have larger spatial dimensions and fewer channels (e.g. 56$\times$56$\times$48), which renders them memory-bound. Highly optimized convolution is more appropriate for these layers. 
Guided by this principle, we choose a pure convolutional network as our base architecture, specifically ConvNeXt~\citep{liu2022convnet} which absorbed several key components from ViTs and competes favorably against ViTs. We gradually ``lighten'' the network to achieve a more favorable balance between latency and accuracy.
For speed metric, we utilize on-device latency, measured on an actual iPhone 13 and compiled by Core ML Tools~\citep{coreml}, rather than FLOPs and parameter counts in
previous methods~\citep{mehta2021mobilevit, chen2022mobile, zhang2022edgeformer}, which are not well correlated with latency. Regarding performance, we follow
the training recipe in ConvNeXt while removing the layer scale to align prior methods~\citep{li2022efficientformer, wang2024repvit} for a fair comparison. Please refer to Sec.~\ref{app:experimental setting} in the supplementary material for more details. 
To initiate our study, we systematically scale down the ConvNeXt by reducing the number of blocks and the width. This results in a lightweight model with a latency of 1.07 ms and a Top-1 accuracy of 74.9\%, serving as our initial baseline.
\subsection{Lightening Baseline} \label{sec:macro_micro}
\begin{figure*}[t]
	\centering
	\includegraphics[width=1\linewidth]{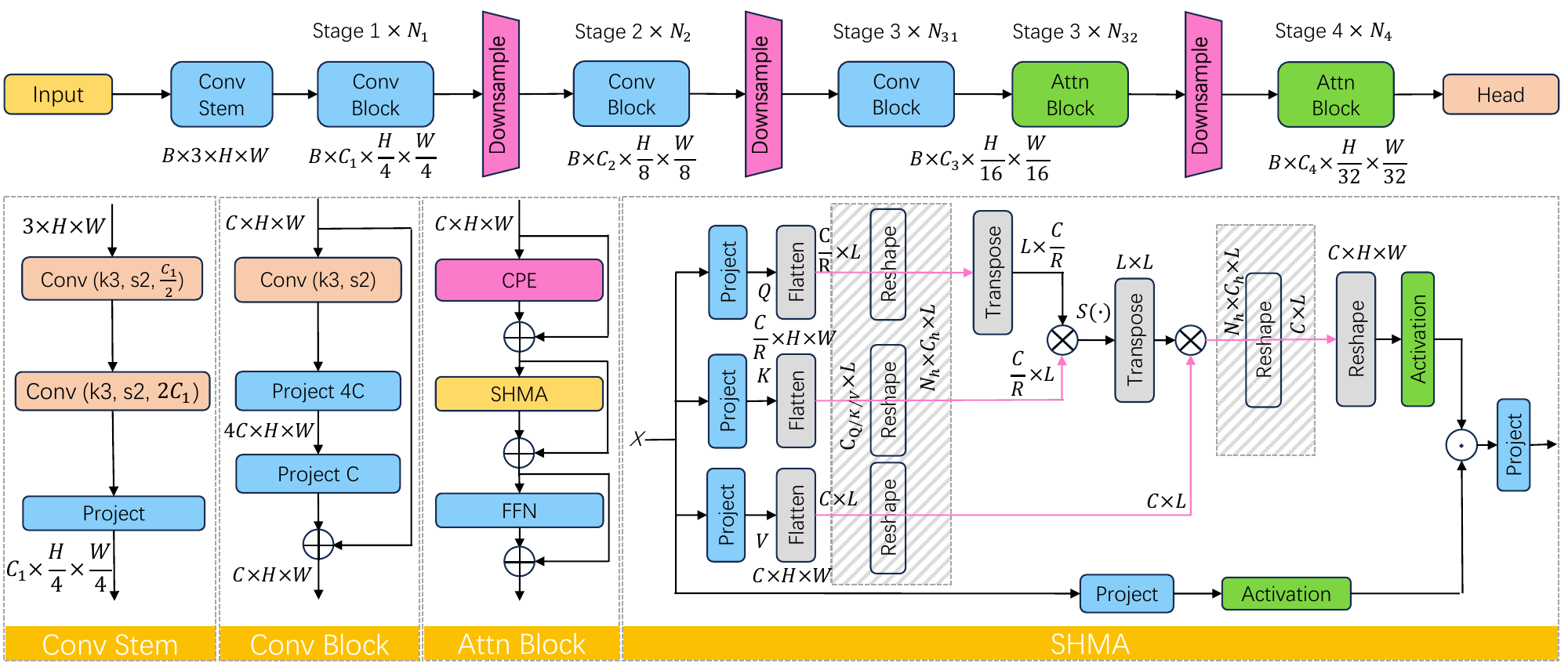}
	\vspace{-2mm}
	\caption{\textbf{Overview of iFormer architecture, detailed convolutional stem, block design, and SHMA.} The hatched area in SHMA indicates extra memory-intensive reshaping operations that are eliminated by SHMA. $S(\cdot)$ denotes the softmax function. $R$ is the ratio for reducing channels of query and key. It is set to 2 in iFormer. We omit BN following project or convolution for simplicity.}
	\label{fig:arch}
	\vspace{-6mm}
\end{figure*}

\paragraph{Seeing Better with Early Convolutions}
Following ViTs, ConvNeXt adopts an aggressive ``patchify" strategy as the stem cell, specifically by splitting the input image into a series of non-overlapping patches via a 4x4 non-overlapping convolutional layer. However, some studies~\citep{xiao2021early,chen2022mixformer} indicate that an early convolutional stem can increase optimization stability and facilitate faster model convergence. Moreover, compared to general models, lightweight models typically have fewer parameters and a reduced capacity. An aggressive non-overlapping layer may lead to the premature loss of rich information. Consequently, we opt to replace the non-overlapping ``patchify" stem with a 
stack of overlapping convolutional layers, as shown in Fig.~\ref{fig:arch}. This modification elevates the top-1 accuracy to 76.7\% with a neglectable increase in latency of 0.1 ms.
\paragraph{Normalization} \label{sec:norm}
An obvious difference between ConvNeXt and previous CNNs is the normalization layer. ConvNeXt utilizes Layer Normalization (LN)~\citep{ba2016layer}, commonly used in Natural Language
Processing (NLP), whereas the latter uses Batch Normalization (BN)~\citep{ioffe2015batch}. Albeit its superior performance, LN requires on-the-fly statistics calculation in inference along with division and square root operations, leading to inefficiency on mobile hardware~\citep{yang2022unified}. On the contrary, BN operates with fixed statistics during inference as an offline method and can be seamlessly fused with other linear operations, providing a ``free lunch". This significantly reduces computational demands and memory overheads on mobile devices. Therefore, we substitute LN with BN throughout the network and merge it during inference. Additionally, we also substitute non-overlapping downsample layers with overlapping counterparts. These adjustments result in a reduction of overall latency to 0.77 ms while enhancing the Top-1 accuracy slightly to 77.10\%.
\paragraph{Going Deeper}
There is considerable evidence indicating that increasing the depth of a model can enhance its capacity and yield performance benefits~\citep{touvron2021going, yang2022focal}. Most lightweight models typically stack more blocks to boost performance within constrained resources, as exemplified by the MobileNet series~\citep{howard2019searching, qin2024mobilenetv4}. In this study, we explore the potential of deepening ConvNeXt by increasing the number of blocks in each stage from (2,2,6,2) to (3,3,9,3). This increase in depth leads to a substantial improvement, raising the accuracy from 77.1\% to 78.8\%, although causing a temporary increase in latency to 1.05 ms.
\paragraph{Stage Ratio}
The stage ratio in ConNeXt is not optimized for lightweight models. A substantial number of depthwise convolutions in the early stages incurs significant memory transfer costs. Meanwhile, the presence of many blocks with a channel expansion ratio of 4 in the Feed-Forward Network (FFN) in the last stage, which already has a high channel dimension, imposes substantial computational demands. These factors lead to a sub-optimal allocation of computational resources. To address these issues, we propose reallocating more computational resources to the third stage while reducing memory access costs in the early stage. Specifically, the blocks in each stage are adjusted from (3,3,9,3) to (2,2,18,2). As expected,
this achieves a better balance between latency and performance, with Top-1 accuracy increasing to 79.1\% while enjoying a lower latency of 1.01ms.

\begin{wraptable}{r}{5cm}
	\caption{\textbf{Latency under different convolutional kernel sizes.}}
	\label{tab:kernel size}
	\centering
	\vspace{-2mm}
	\small
	\scalebox{0.95}{
		\begin{tabular}{cc}
			\toprule
			Kernel Size & Latency (ms) \\
			\midrule
			3$\times$3 & 1.00 \\
			7$\times$7 & 1.01 \\
			\bottomrule
		\end{tabular}
	}
	\vspace{-2mm}
\end{wraptable}
\paragraph{Kernel Size}
Here we examine the effects of different kernel sizes in mobile settings and observe that utilizing a larger kernel size introduces nearly no latency burden, as shown in Table~\ref{tab:kernel size}. So we maintain the convolutional kernel size at 7$\times$7 in each basic block, consistent with ConvNeXt. Furthermore, previous approaches use a kernel size of 3$\times$3 in the convolutional stem. This small receptive field may hinder feature representation during the early downsampling process. As previously noted, the early layers are memory-bound, allowing for opportunities to employ compute-intensive operations (\textit{i.e.}, dense convolution). Therefore, we enlarge the kernel size of the dense convolutional layer in the stem cell to 5$\times$5. As illustrated in Fig.~\ref{fig:roadmap}, this change has no impact on inference latency while enhancing Top-1 accuracy by 0.3\%.
\subsection{Single-Head Modulation Attention} \label{sec:shma}
\paragraph{Single-Head \textit{vs.} Multi-Head}
ViTs typically apply MHA, which projects the queries, keys, and values multiple times with different learnable linear projections and performs multiple attention functions simultaneously. In practice, the multi-head mechanism requires the reshaping of feature maps first, causing large memory access and transfer costs. This can seriously impact inference latency, especially on resource-constrained mobile devices. To investigate this issue, we substitute the last half of the convolutional blocks in the third stage and all blocks in the last stage with standard ViT blocks, as depicted in Fig.~\ref{fig:arch}. 
We refer to this hybrid network as the MHA baseline. Next, we build another network by substituting the MHA with Single-Head self-Attention (SHA), referring to it as the SHA baseline. The comparison is shown in Table~\ref{tab:single vs. multi head}. The SHA baseline shows a 1.25$\times$ acceleration over its MHA counterpart on the iPhone 13. This verifies that additional reshaping operations in MHA incur significant memory access costs, leading to a considerable decline in inference speed.

This naturally calls for optimizing MHA. Recent methods~\citep{pan2022edgevits, qin2024mobilenetv4} primarily focus on downsampling the query or the key, which may hurt global attention capacity. Instead, we aim to reduce the redundant reshaping of MHA while preserving all token-to-token interactions. Previous works~\citep{michel2019sixteen, yun2024shvit} indicate that a single attention head can approach the performance of multiple heads in general plain transformer models, such as DeiT. To investigate this on the mobile application, we analyze the average cosine similarity of multiple heads within the same layer of the aforementioned MHA baseline, which is a hierarchical lightweight network, and present our findings in Fig.~\ref{fig:similarity}. We clearly see that the average cosine similarity reaches 50\% and even 75\% in the final layer. Furthermore, the SHA baseline, as shown in Table~\ref{tab:single vs. multi head}, exhibits only a negligible accuracy drop of 0.1\%. These suggest that SHA achieves a more favorable balance between accuracy and latency, obtaining an accuracy of 79.8\% with a latency of 1.12 ms.
\paragraph{Modulation Attention} \label{med:ma}
We further introduce a novel modulation attention to boost performance and strengthen flexibility in modeling, as illustrated in Fig.~\ref{fig:arch}. Formally, we start from the abstracted modulation mechanism~\citep{ma2024efficient}, similar to the gate mechanism~\cite{shazeer2020glu}. Assume we are given an input feature map $\ \mathbf{x}\in \mathbb{R}^{C\times H\times W}$ where $C$, $H$, and $W$ denote the channels, height, and width of the feature map. The modulated output can be written as follows:
\begin{align}
	\begin{split}
		\mathbf{x_o}& =f(\mathbf{x})  \odot  \text{ctx}(\mathbf{x}),
		\label{eq:modualtion}
	\end{split}
\end{align}
where $f(\cdot)$ denotes the feature mapping branch and $\text{ctx}(\cdot)$ is the context modeling branch. The output $\mathbf{x_o}$ is the fused features from both branches via efficient element-wise multiplication. The key idea of our approach is to modulate the feature using SHA instead of relying on convolutional layers, as seen in previous works~\citep{yang2022focal, ma2024efficient}. Since SHA captures global interactions through self-attention, it excels in extracting rich contextual information and better controlling the flow of information. This process can be expressed as follows:
\begin{align}
	\begin{split}
	\text{ctx}(\mathbf{x}) & = \text{SHA}(\mathbf{W^Q}\mathbf{x}, \mathbf{W^K}\mathbf{x}, \mathbf{W^V}\mathbf{x}), 
	\end{split}
\end{align}
where $\mathbf{W^Q}$, $\mathbf{W^K}$, $\mathbf{W^V}$ are the project weights for query, key, and value, respectively. For simplicity, we omit the bias term. To minimize inference costs, we utilize a single projection layer in the feature mapping branch. To enhance expressivity and improve optimization stability, we apply individual nonlinear activation functions to both branches, as follows:
\begin{align}
	\begin{split}
		\mathbf{x_o} = \sigma({\mathbf{W^M}\mathbf{x}})  \odot \sigma(\text{ctx}(\mathbf{x})) ,
	\end{split}
\end{align}
where $\sigma$ is the sigmoid function and $\mathbf{W^M}$ denotes the feature projection weight. We also experiment with various activation functions for modulation in Sec.~\ref{abl:act} and observe that the sigmoid works rather well.
Finally, the output from the modulation attention is projected in a manner as standard attention.

Equipped with Single-Head Modulation Attention (SHMA), our model improves the accuracy to 80.4\% with an intermediate latency of 1.15 ms. This performance notably surpasses that of the recent MobileNetV4, which achieves an accuracy of 79.9\%.
\paragraph{Reducing Width} \label{sec:rw}
Until now, we have developed a lightweight network that performs pretty well, but at a bit slow speed. To push the trade-off toward the state-of-the-art, we revise the width configuration in the SHMA. The modulation mechanism enriches the output by enabling more dynamic modeling in both spatial and channel dimensions, making it possible to use a weaker SHA and FFN. In light of this, we reduce the head dimension in the SHMA (\textit{i.e.}, $\mathbf{W^Q},  \mathbf{W^K}$) to a small factor of the feature dimension, further details can be found in Table~\ref{tab:arch spec} in the supplementary material. Simultaneously, we shrink the expansion ratio in FFN following SHMA from 4 to 3. This process obtains a lower latency of 1.10 ms, although a slight drop of 0.2\% in accuracy.
\paragraph{Positional Embedding}
Last but not least, positional information plays a crucial role in self-attention as it regards input as a set of tokens. Adding positional embedding will help the attention learn permutation-variant features. We apply conditional positional encodings (CPE)~\citep{chu2021conditional} that are dynamically generated and conditioned on the local neighborhood of the input tokens, as illustrated in Fig.~\ref{fig:arch}.
The integration of CPE further enhances our model's performance, achieving a Top-1 accuracy of 80.4\% with only 1.10 ms, establishing a state-of-the-art trade-off.
\paragraph{iFormer} \label{sec:iformer}
The result of these modifications is an extremely fast and efficient hybrid network, which we denote \textit{iFormer}. The overall architecture is depicted inFig.~\ref{fig:arch}. It integrates fast local convolutional layers in the early stages that operate on higher resolution and global SHMA in later stages which processes lower resolution. Besides, we create a series of iFormer models tailored to various hardware resource constraints. For detailed architectural hyperparameters of these model variants, please refer to Table~\ref{tab:arch spec} in the supplementary material.
\section{Experiments}
\begin{table}[t]
	\caption{{\textbf{Classification results on ImageNet-1K.} $^{\dag}$ indicates models that are trained with a variety of advanced training strategies including complex reparameterization, distillation, optimizer, and so on. We provide a more comprehensive comparison in Sec.~\ref{app:comprehensive comparison} in the supplementary material.}}
	\label{tab:image classification}
	\centering
	\vspace{-2mm}
	\small
	\scalebox{0.9}{
		\begin{tabular}{cccccccccc}
			\toprule
			\multirow{2}{*}{Model}   & \multirow{2}{*}{Params (M)} & \multirow{2}{*}{GMACs} &  \multirow{2}{*}{\makecell{Latency $\downarrow$ \\ (ms)}}  & \multirow{2}{*}{Reso.}   & \multirow{2}{*}{Epochs} & \multirow{2}{*}{Top-1 (\%)}  \\
			\\
			\midrule
			MobileNetV2 1.0x~\citeyearpar{sandler2018mobilenetv2} & 3.4 & 0.30 & 0.73 & 224 & 500 & 72.0 \\
			MobileNetV3-Large 0.75x~\citeyearpar{howard2019searching} & 4.0 & 0.16 & 0.67 & 224 & 600 & 73.3 \\
			MNV4-Conv-S~\citeyearpar{qin2024mobilenetv4} & 3.8 & 0.20 & 0.60 & 224 & 500 & 73.8 \\
			\rowcolor[gray]{0.92}
			\textbf{iFormer-T} & 2.9& 0.53 & \textbf{0.60} &  224  & 300  & \textbf{74.1}\\
			\midrule
			MobileNetV2 1.4x~\citeyearpar{sandler2018mobilenetv2} & 6.9 & 0.59 & 1.02 & 224 & 500 & 74.7 \\
			MobileNetV3-Large 1.0x~\citeyearpar{howard2019searching} & 5.4 & 0.22 & 0.76 & 224 & 600 & 75.2 \\
			SwiftFormer-XS~\citeyearpar{shaker2023swiftformer}&3.5 & 0.60 & 0.95 & 224 & 300 & 75.7 \\
			SBCFormer-XS~\citeyearpar{lu2024sbcformer} & 5.6 & 0.70 & 0.79 & 224 & 300 & 75.8  \\
			${\text{GhostNetV3 1.0x}^{\dag}}$~\citeyearpar{liu2024ghostnetv3} & 6.1 & 0.17 & 0.99 &224 & 600 & 77.1 \\
			MobileOne-S2~\citeyearpar{vasu2023mobileone} & 7.8 & 1.30 & 0.92 & 224 & 300 & 77.4 \\
			RepViT-M1.0~\citeyearpar{wang2024repvit}  & 6.8 & 1.10 & 0.85 & 224 & 300 & 78.6 \\
			\rowcolor[gray]{0.92}
			\textbf{iFormer-S} & 6.5 & 1.09 & \textbf{0.85} &  224  & 300  & \textbf{78.8}\\
			\midrule
			EfficientMod-xxs~\citeyearpar{ma2024efficient} & 4.7 & 0.60 & 1.29 & 224 & 300 & 76.0 \\
			SBCFormer-S~\citeyearpar{lu2024sbcformer} & 8.5 & 0.90 & 1.02 & 224 & 300 & 77.7 \\
			MobileOne-S3~\citeyearpar{vasu2023mobileone} & 10.1 & 1.90 & 1.16 & 224 & 300 & 78.1  \\
			SwiftFormer-S~\citeyearpar{shaker2023swiftformer}&6.1 & 1.00 & 1.12 & 224 & 300 & 78.5 \\
			${\text{GhostNetV3 1.3x}^{\dag}}$~\citeyearpar{liu2024ghostnetv3}  & 8.9 & 0.27 & 1.24 & 224 & 600 & 79.1 \\
			FastViT-T12~\citeyearpar{vasu2023fastvit} & 6.8 & 1.40 & 1.12 & 256 & 300 & 79.1 \\
			RepViT-M1.1~\citeyearpar{wang2024repvit}  & 8.2 & 1.30 & 1.04 & 224 & 300 & 79.4 \\
			MNV4-Conv-M~\citeyearpar{qin2024mobilenetv4} & 9.2 &1.00 & 1.08 & 256 & 500 & 79.9 \\
			\rowcolor[gray]{0.92}
			\textbf{iFormer-M} & 8.9 & 1.64 & \textbf{1.10} &  224  & 300  & \textbf{80.4} \\
			\midrule
			Mobile-Former-294M~\citeyearpar{chen2022mobile} &11.4 & 0.29 & 2.66 & 224 & 450 & 77.9 \\
			MobileViT-S~\citeyearpar{mehta2021mobilevit}& 5.6 & 2.00 & 3.55 & 256 & 300 & 78.4 \\
			MobileOne-S4~\citeyearpar{vasu2023mobileone} & 14.8 & 2.98 & 1.74 & 224 & 300 & 79.4  \\
			SBCFormer-B~\citeyearpar{lu2024sbcformer} & 13.8 & 1.60 & 1.44 & 224 & 300 & 80.0 \\
			${\text{GhostNetV3 1.6x}^{\dag}}$~\citeyearpar{liu2024ghostnetv3} & 12.3 & 0.40 & 1.49 & 224 & 600 & 80.4 \\
			EfficientViT-B1-r288~\citeyearpar{cai2023efficientvit} & 9.1 & 0.86 & 3.87 & 288 & 450 & 80.4 \\
			FastViT-SA12~\citeyearpar{vasu2023fastvit} & 10.9 & 1.90 & 1.50 & 256 & 300 & 80.6 \\
			MNV4-Hybrid-M~\citeyearpar{qin2024mobilenetv4} & 10.5  & 1.20 & 1.75 & 256 & 500 & 80.7 \\
			SwiftFormer-L1~\citeyearpar{shaker2023swiftformer}& 12.1 & 1.60 & 1.60 & 224 & 300 & 80.9 \\
			EfficientMod-s~\citeyearpar{ma2024efficient} & 12.9 & 1.40 & 2.57 & 224 &300 & 81.0 \\
			RepViT-M1.5~\citeyearpar{wang2024repvit}  & 14.0 & 2.30& 1.54 & 224 & 300 & 81.2 \\
			\rowcolor[gray]{0.92}
			\textbf{iFormer-L} & 14.7 & 2.63 & \textbf{1.60} &  224  & 300  & \textbf{81.9}\\
			\bottomrule
		\end{tabular}
	}
	\vspace{-2mm}
\end{table}
\subsection{Image Classification}
\paragraph{Settings.} We first evaluate our models on classification on ImageNet-1K~\citep{deng2009imagenet}. To ensure a fair comparison with prior studies, we follow the previous training recipe~\citep{touvron2021training, liu2022convnet} and train all models for 300 epochs with a standard image size of 224x224. Please refer to Sec.~\ref{app:experimental setting} in the supplementary material for details. Besides Top-1 validation accuracy, we also report the latency measured on an iPhone 13 with models compiled by Core ML Tools~\citep{coreml} under a batch size of 1, as done in~\citep{li2023rethinking, wang2024repvit, vasu2023mobileone}. It's worth highlighting that we do not apply any advanced strategies such as distillation~\citep{li2023rethinking} and reparameterization~\citep{ding2021repvgg}.

\begin{wraptable}{r}{9cm}
	\caption{\textbf{Results with distillation on ImageNet-1K.} * indicates the model is trained with a strong training strategy (\textit{i.e.}, reparameterization).}
	\label{tab:distillation}
	\centering
	\vspace{-2mm}
	\small
	\scalebox{0.8}{
		\begin{tabular}{cccccccccc}
			\toprule
			\multirow{1}{*}{Model}          &  \multirow{1}{*}{\makecell{Latency (ms)}} & \multirow{1}{*}{Reso.}  & \multirow{1}{*}{Epochs} & \multirow{1}{*}{Top-1 (\%)} \\
			\midrule
			EfficientFormerV2-S1~\citeyearpar{li2023rethinking} & 1.02 & 224 & 300 & 79.0 \\
			EfficientFormerV2-S1~\citeyearpar{li2023rethinking} & 1.02 & 224 & 450 & 79.7 \\
			MobileViGv2-S*\citeyearpar{avery2024scaling} &1.24 & 224 & 300 & 79.8 \\
			FastViT-T12*~\citeyearpar{vasu2023fastvit} & 1.12 & 256 & 300 & 80.3 \\
			RepViT-M1.1*~\citeyearpar{wang2024repvit}  & 1.04 & 224 & 300 & 80.7 \\
			\rowcolor[gray]{0.92}
			\textbf{iFormer-M} & \textbf{1.10} &  224  & 300  & \textbf{81.1}\\
			\midrule
			SHViT-S4~\citeyearpar{yun2024shvit} & 1.48 & 224 & 300 & 80.2 \\
			EfficientFormerV2-S2~\citeyearpar{li2023rethinking} & 1.60 & 224 & 300 & 81.6 \\
			MobileViGv2-M\citeyearpar{avery2024scaling} &1.70 & 224 & 300 & 81.7 \\
			FastViT-SA12*~\citeyearpar{vasu2023fastvit} & 1.50 & 256 & 300 & 81.9 \\
			EfficientFormerV2-S2~\citeyearpar{li2023rethinking} & 1.60 & 224 & 450 & 82.0 \\
			RepViT-M1.5*~\citeyearpar{wang2024repvit}  & 1.54  & 224 & 300 & 82.3 \\
			\rowcolor[gray]{0.92}
			\textbf{iFormer-L} & \textbf{1.60} &  224  & 300  & \textbf{82.7}\\
			\bottomrule
		\end{tabular}
	}
	\vspace{-2mm}
\end{wraptable}

Table~\ref{tab:image classification} summarizes a comparison between our iFormer and state-of-the-art lightweight models, organized by latency. iFormer demonstrates a Pareto-optimal trade-off between accuracy and latency. For example, iFormer-M obtains 80.4\% top-1 accuracy with a latency of only 1.1 ms, surpassing recent MobileNetV4-Conv-M and RepViT-M1 by 0.5\% and 1.0\%, respectively. This is noteworthy considering that MobileNetV4 requires a longer training schedule (500 \textit{vs.} 300) and takes a larger input resolution (256 \textit{vs.} 224). When compared to other recent models using reparameterization, including FastViT-T12, GhostNetV3-1.3$\times$, and MobileOne-S3, iFormer-M achieves superior accuracy while maintaining lower latency. Moreover, iFormer outperforms various hybrid networks. Thanks to the efficient SHMA, iFormer-L achieves more outstanding performance than other attention variants, such as multi-query attention in MNV4-Hybrid-M, additive attention in SwiftFormer-L1, and linear attention in EfficientVIT-B1-r288. 
\paragraph{Results with distillation on ImageNet-1K.}
We conducted rigorously fair training as the previous methods above. Recently, some works report enhanced performance leveraging more advanced training strategies. 
We investigate whether these training recipes can also improve iFormer. Following previous works~\citep{li2023rethinking, wang2024repvit}, we employ the RegNetY-16GF~\citep{radosavovic2020designing} model with a top-1 accuracy of 82.9\% as the teacher model for distillation. Our findings reveal that iFormer improves obviously over its counterpart without distillation. For example, iFormer-L shows a 1.0\% increase under the same latency. iFormer also outperforms EfficientFormerV2-S2, despite the latter being trained with a 1.5$\times$ longer schedule.
\subsection{Object Detection and Instance Segmentation}
To validate the effectiveness of iFormer on downstream tasks, we train Mask R-CNN~\citep{he2017mask} with iFormer as the backbone for 12 epochs (1$\times$), using the MMDetection toolkit~\citep{chen2019mmdetection}. We also report backbone latency measured at a resolution of 512$\times$512 on an iPhone 13. The results are presented in Table~\ref{tab:coco}. In comparison to lightweight models, iFormer-M surpasses FastViT-SA12 by +1.9\%/+2.0\% in AP$^\text{box}$ /AP$^\text{mask}$ while running 1.32$\times$ faster. iFormer-L also obtains +0.1\%/+0.6\% in AP$^\text{box}$ /AP$^\text{mask}$ than EfficientMod-S, which utilizes a convolutional modulation mechanism to learn dynamics similar to self-attention. Notably, EfficientMod-S operates 3.7$\times$ slower when processing high-resolution input, underscoring that the proposed novel attention mechanism is more suitable for mobile networks. Meanwhile, when compared to general networks that are not optimized for mobile applications, iFormer demonstrates significant advantages. For instance, iFormer-L exceeds the performance of ConvNeXt-T with improvements of +1.2\%/+1.4\% in AP$^\text{box}$ /AP$^\text{mask}$, while requiring fewer parameters and only 50\% mobile latency, suggesting iFormer's efficient design in feature extraction and strong potential for mobile applications.
\begin{table}[t]
	\centering
	\vspace{-2mm}
	\small
	\caption{
		\textbf{Object detection \& instance segmentation} results on MS COCO 2017 using Mask R-CNN.
		\textbf{Semantic segmentation} results on ADE20K using the Semantic FPN framework. We measure all backbone latencies with image crops of 512$\times$512 on iPhone 13 by Core ML Tools. Failed indicated that the model runs too long to report latency by the Core ML.
	}
	\resizebox{0.95\linewidth}{!}{
		\begin{tabular}{c|c|c|ccc|ccc|c}
			\toprule
			\multirow{2}{*}{Backbone} &\multirow{2}{*}{\makecell{Param \\ (M)}} & \multirow{2}{*}{\makecell{Latency $\downarrow$ \\ (ms)}} & \multicolumn{3}{c|}{Object Detection} & \multicolumn{3}{c|}{Instance Segmentation} & \multicolumn{1}{c}{Semantic}  \\ 
			\cmidrule{4-10} &  &  & AP$^\text{box}$    & AP$^\text{box}_{50}$   & AP$^\text{box}_{75}$   & AP$^\text{mask}$    & AP$^\text{mask}_{50}$   & AP$^\text{mask}_{75}$   & mIoU   \\
			\hline
			\midrule
			EfficientNet-B0~\citeyearpar{tan2019efficientnet}	& 5.3 & 4.55 & 31.9 & 51.0 & 34.5 & 29.4 & 47.9 & 31.2 & - \\
			ResNet18~\citeyearpar{he2016deep} & 11.7   &  2.85     & 34.0   & 54.0    & 36.7    & 31.2   & 51.0    & 32.7   & 32.9   \\
			PoolFormer-S12~\citeyearpar{yu2022metaformer}     & 11.9     &   5.70 & 37.3   & 59.0    & 40.1    & 34.6   & 55.8    & 36.9     & 37.2  \\
			EfficientFormer-L1~\citeyearpar{li2022efficientformer}     & 12.3 & 3.50 & 37.9 & 60.3 & 41.0 & 35.4 & 57.3 & 37.3 & 38.9  \\
			FastViT-SA12~\citeyearpar{vasu2023fastvit}     & 10.9 & 5.27 & 38.9 & 60.5 & 42.2 & 35.9 & 57.6 & 38.1 & 38.0  \\
			RepViT-M1.1~\citeyearpar{wang2024repvit}    & 8.2 & 3.18 & 39.8 & 61.9 & 43.5 & 37.2 & 58.8 & 40.1 & 40.6 \\
			\rowcolor[gray]{0.92}
			iFormer-M   &     \text{8.9}  &   \text{4.00}              &  \text{40.8}  &  \text{62.5}   & \text{44.8}  &    \text{37.9}    &   \text{59.7}      &  \text{40.7}   & \text{42.4}  \\
			\midrule
			ResNet50~\citeyearpar{he2016deep}  & 25.5 & 7.20 & 38.0 & 58.6 & 41.4 & 34.4 & 55.1 & 36.7 & 36.7 \\
			PoolFormer-S24~\citeyearpar{yu2022metaformer}  &21.4    &     10.0    & 40.1   & 62.2    & 43.4    & 37.0   & 59.1    & 39.6  &  40.3 \\
			ConvNeXt-T~\citep{liu2022convnet} & 29.0 & 13.6 & 41.0 & 62.1 & 45.3 & 37.7 & 59.3 & 40.4 & 41.4 \\
			EfficientFormer-L3~\citeyearpar{li2022efficientformer}  & 31.3 & 8.40 & 41.4 & 63.9 & 44.7 & 38.1 & 61.0 & 40.4 & 43.5 \\
			RepViT-M1.5~\citeyearpar{wang2024repvit} & 14.0 & 5.00 & 41.6 & 63.2 & 45.3 & 38.6 & 60.5 & 41.5 & 43.6 \\
			PVTv2-B1~\citeyearpar{wang2022pvt} & 14.0 & 27.00 & 41.8 & 64.3 & 45.9 & 38.8 & 61.2 & 41.6 & 42.5 \\
			FastViT-SA24~\citeyearpar{vasu2023fastvit} & 20.6 & 8.97 & 42.0 & 63.5 & 45.8 & 38.0 & 60.5 & 40.5 & 41.0 \\
			EfficientMod-S~\citeyearpar{ma2024efficient} & 32.6 & 24.30 & 42.1 & 63.6 & 45.9 & 38.5 & 60.8 & 41.2 & 43.5 \\
			Swin-T~\citeyearpar{liu2021swin} & 28.3 & Failed & 42.2 & 64.4 & 46.2 & 39.1 & 61.6 & 42.0 & 41.5 \\
			\rowcolor[gray]{0.92}
			iFormer-L    &  \text{14.7} &   \text{6.60}      & \text{42.2}   & \text{64.2}    &  \text{46.0}   &  \text{39.1}   & \text{61.4}   & \text{41.9}     &   \text{44.5}   \\
			\bottomrule
		\end{tabular}
	}
	\vspace{-3mm}
	\label{tab:coco}
\end{table}
\subsection{Semantic Segmentation}
We conduct experiments on the ADE20K~\citep{zhou2017scene} using the Semantic FPN~\citep{kirillov2019panoptic}, based on the MMSegmentation toolkit~\citep{mmseg2020}. Thanks to its efficient attention design, iFormer outperforms all competing methods in mIoU with similar and much lower latency. For example, iFormer-L surpasses FastViT-SA24 by +3.5\% in mIoU with a 1.36$\times$ faster inference speed. In addition, iFormer-M demonstrates superior mIoU compared to general networks, which typically exhibit substantially greater latency when processing higher-resolution inputs on mobile devices. Although PVTv2-B utilizes downsampled attention, it still requires 27 ms for latency. Similarly, Swin-T involves intensive operations in window partitioning, making it less suitable for mobile applications. Running at 6.6 ms, iFormer-L achieves +2.0\% better mIoU than PVTv2-B1 and +3.0\% better than Swin-T. 
These results suggest that the proposed attention mechanism offers significant benefits for tasks requiring the perception of fine-grained details.
\section{Ablation Studies}
\paragraph{Activation Function} \label{abl:act}
\begin{wraptable}{r}{8.7cm}
	\caption{\textbf{Activation function comparison in SHMA.} Post-BN indicates that BN is applied after projection. Pre-LN means that LN is implemented before the projection, as in standard MHA~\citep{vaswani2017attention}.}
	\label{tab:act comparison}
	\centering
	\vspace{-2mm}
	\small
	\setlength{\tabcolsep}{2.5pt}
	\scalebox{0.9}{
		\begin{tabular}{ccccc}
			\toprule
			SHMA Setting & Params (M) & GMACs & Latency (ms) & Top-1 Acc. (\%)\\
			\midrule
			SiLU + Post-BN & 8.9 & 1.60& 1.10ms & Diverged \\
			SiLU + Pre-LN & 8.9 & 1.64& 1.17ms & 80.3 \\
			Sigmoid + Post-BN & 8.9 & 1.60 & 1.10ms & 80.4\\
			\bottomrule
		\end{tabular}
	}
	\vspace{-2mm}
\end{wraptable}
Here we explore whether an activation function without an upper bound can enhance the SHMA by allowing neurons to express arbitrarily large values. We compare the widely used Sigmoid Linear Unit (SiLU)~\citep{shazeer2020glu} with the sigmoid function and present the results in Table~\ref{tab:act comparison}. Directly replacing the activation function in SHMA with SiLU will encounter diverging loss during training. The underlying cause is primarily attributed to the element-wise multiplication of the unbounded context branch. To address this, we replace Post-BN in SHMA with Pre-LN, as LN adaptively normalizes each token feature. The modified model experiences a slight decrease in accuracy but incurs an additional 0.07 ms latency, primarily brought by LN. The results suggest that the sigmoid function not only mitigates training instability but also facilitates better convergence. 
\paragraph{Choice of Conv v.s. ViT Blcoks}
In Section~\ref{sec:shma}, we replace the convolutional blocks in Stages 3 and 4 with the proposed SHMA block. We provide further ablation studies on the choice of ratio for the ViT blocks. Specifically, We choose the model after enlarging the kernel size as a starting point, then we progressively replace the convolutional blocks in Stages 3 and 4. We do not modify Stages 1 and 2 as their larger spatial dimensions would considerably increase the memory requirements for the self-attention mechanism.
\begin{table}[h]
	\caption{\textbf{Different ratio of ViT Block.}}
	\label{tab:ratio of vit blocks}
	\centering
	\vspace{-3mm}
	\small
	\setlength{\tabcolsep}{1.5pt}
	\scalebox{0.87}{
		\begin{tabular}{lcccc}
			\toprule
			Ratio Setting & Params (M) & GMACs & Latency (ms) & Top-1 Acc. (\%)\\
			\midrule
			Baseline                                                         & 9.4M  &1760M &  1.0ms & 79.4 \\
			Replacing 22\% Conv Blocks in Stage 3 as SHA                      & 9.1M  &1724M & 1.02ms &  79.5 \\
			Replacing 22\% Conv Blocks in Stage 3 as SHMA                     & 9.2M  &1739M & 1.04ms & 79.6 \\
			Replacing 50\% Conv Blocks in Stage 3 as SHA                      & 8.8M  &1689M & 1.04ms &   79.5 \\
			Replacing 50\% Conv Blocks in Stage 3 as SHMA                     & 8.9M  &1712M & 1.07ms & 79.8 \\
			Replacing 78\% Conv Blocks in Stage 3 as SHA                      & 8.3M  &1635M & 1.12ms &   79.3 \\
			Replacing 78\% Conv Blocks in Stage 3 as SHMA                     & 8.5M  &1685M & 1.17ms & 79.6 \\
			Replacing 100\% Conv Blocks in Stage 3 as SHA                     & 7.9M  &1599M & 1.17ms &   78.1 \\
			Replacing 100\% Conv Blocks in Stage 3 as SHMA                    & 8.3M  &1665M & 1.25ms & 79.0 \\
			\midrule
			Replacing 100\% Conv Blocks in Stage 3 as SHMA and 100\% in Stage 4 &10.0M  &1792M & 1.15ms & 80.4 \\
			\bottomrule
		\end{tabular}
	}
	\vspace{-3mm}
\end{table}

We present our findings in Table~\ref{tab:ratio of vit blocks}. Given that Stage 4 contains only two blocks, we do not conduct further splitting for the ratio. As illustrated in Table~\ref{tab:ratio of vit blocks}, although the ViT block has lower FLOPs, it still incurs increased runtime. Substituting all the convolutional blocks in Stage 3 results in the worst performance and the highest latency. Instead, by replacing half of the convolutional blocks in the third stage and all blocks in the final stage, we can better integrate these two operators, thus achieving a favorable trade-off between accuracy and latency.

\paragraph{Scaling to Larger Model}
Although iFormer is designed for mobile-device applications, the combination of fast local representation capacity of convolution and the efficient global modeling proficiency of the proposed SHMA enables its scalability for a broader range of applications.
To demonstrate the scalability of iFormer, we developed a larger model named iFormer-H with 99M parameters and trained it for 300 epochs following the same strategy outlined in Section~\ref{app:experimental setting}. It is important to note that we add drop path and layer scale, which are commonly used in the training of larger models~\citep{liu2022convnet, tu2022maxvit, shi2024transnext}. 

\begin{wraptable}{r}{9.5cm}
	\caption{\textbf{Scaling to the larger model with 99M parameters.}}
	\label{tab:scale}
	\centering
	\small
	\vspace{-3mm}
	\setlength{\tabcolsep}{1.5pt}
	\scalebox{0.98}{
		\begin{tabular}{ccccc}
			\toprule
			Model & Params (M) & GMACs (G) & Top-1 Acc. (\%)\\
			\midrule
			ConvNeXt-Base~\citeyearpar{liu2022convnet} & 89 & 15.4 & 83.8  \\
			TransNeXt-Base~\citeyearpar{shi2024transnext} & 90 & 18.4 & 84.8 \\
			\rowcolor[gray]{0.92}
			iFormer-H & 99 & 15.5 & 84.8 \\
			MaxViT-Base~\citeyearpar{tu2022maxvit} & 120 & 24.0 & 84.9 \\
			\bottomrule
		\end{tabular}
	}
	\vspace{-3mm}
\end{wraptable}
We summarize the results in Table~\ref{tab:scale}. A highlight from the results is that iFormer is not specifically designed or trained for this scale. Despite this, iFormer-H outperforms ConvNeXt, achieving a 1.0\% increase in accuracy while maintaining a similar number of FLOPs. Additionally, it demonstrates comparable performance to TransNeXt-Base, despite utilizing fewer FLOPs. These findings indicate the potential for broader applications of iFormer. We plan to explore larger models suitable for mobile devices in future work.
Further ablation studies can be found in Sec.~\ref{app: more ablation} in the supplementary material.

\section{Conclusion}
This work proposes iFormer, which integrates highly optimized convolutional operations for the early layers alongside a novel and efficient single-head modulation attention for the later layers. iFormer achieves SOTA Pareto-front in terms of Top-1 accuracy and mobile latency. We also validate the effectiveness of iFormer on downstream dense prediction tasks, including COCO object detection, instance segmentation, and ADE20K semantic segmentation. These inspiring results highlight the potential for mobile applications. We hope iFormer can facilitate the application of artificial intelligence on more mobile devices. In future work, we will seek to alleviate inference bottlenecks sociated with high-resolution images. Meanwhile, we plan to optimize iFormer for more hardware platforms, such as Android devices and NVIDIA Jetson Nano.
%

\bibliography{iclr2025_conference}
\bibliographystyle{iclr2025_conference}

\clearpage
\appendix
\section{Appendix}
\section{Experimental Settings}\label{app:experimental setting}
\subsection{Image Classification} \label{app:image cls}
\begin{table}[h]
	\centering
	\caption{\textbf{ImageNet-1K training settings}.}
	\vspace{-2mm}
	\begin{tabular}{@{\hskip -0.05ex}l|c@{\hskip 1ex}c}
		training config & iFormer-T/S/M/L/H \\
		\midrule
		resolution & 224$^2$ \\
		weight init & trunc. normal (0.2) \\
		optimizer & AdamW\\
		base learning rate & 4e-3 (T/S/M/L) 8e-3\\
		weight decay & 0.05 \\
		optimizer momentum & $\beta_1, \beta_2{=}0.9, 0.999$\\
		batch size & 4096  [T/S/M/L] 8192 [H] \\
		training epochs & 300\\
		learning rate schedule & cosine decay \\
		warmup epochs & 20 \\
		warmup schedule & linear \\
		layer-wise lr decay & None \\
		randaugment & (9, 0.5) \\
		mixup  & 0.8 \\
		cutmix & 1.0 \\
		random erasing & 0.25\\
		label smoothing & 0.1  \\
		\midrule
		stochastic depth & 0.0 [T/S/M] 0.1 [L] 0.6 [H] \\
		layer scale & None [T/S/M/L] 1e-6 [H]\\
		head init scale & None \\
		gradient clip & None \\
		exp. mov. avg. (EMA) & None \\
	\end{tabular}
	\vspace{-2mm}
	\label{tab:train_detail}
\end{table}
We mainly follow the training recipe of ConvNeXt, while removing stochastic depth, layer scale, and exponential moving average to ensure a fair comparison with prior works. The models are trained for 300 epochs on 8 NVIDIA GPUs with a total batch size of 4096. We employ the same learning rate across all models. It is possible to further improve performance by adjusting the learning rates for different model variants, which we will explore in the future.

For distillation, we use the RegNetY-16GF model as the teacher model and apply a hard distillation loss, following the approach of DeiT~\citep{touvron2021training}. During inference, the average output of the classification head and the distillation head is used as the final output.
\subsection{Object Detection and Semantic Segmentation}
For object detection experiments, we train MaskR-CNN models on the COCO 2017 dataset for 12 epochs using standard training settings from the MMDetection toolkit.

For semantic segmentation experiments, we train Semantic FPN models on the ADE20K dataset for 40,000 iterations using standard training settings from the MMSegmentation toolkit. The input images are cropped to a resolution of 512$\times$512 during training.

For backbone latency, we keep the same input size as training (\textit{i.e.}, 512$\times$512) and measure the mobile latency on an iPhone 13 compiled by Core ML Tools.
\section{More Ablation Studies} \label{app: more ablation}
\paragraph{Different Ways for Reducing Latency}
Here we provide a comparison of different methods for reducing latency, contrasting them with the approach discussed in  Sec.~\ref{sec:rw}. Specifically, we reduce the baseline latency to similar latency by directly removing blocks, cutting down FFN expansion width, and reducing both attention head dimension and FFN expansion dimension simultaneously. 
From the results in Table~\ref{tab:reduce depth or width}, we observe that the removal of a single block in the final stage can lead to a severe drop in accuracy (-0.7\%), indicating that greater depth enhances the model's capacity. Concurrently reducing all FFN expansion widths causes a non-trivial performance degradation (-0.6\%). 
\begin{wraptable}{r}{9.5cm}
	\caption{\textbf{Different ways for reducing latency.}}
	\label{tab:reduce depth or width}
	\centering
	\vspace{-2mm}
	\small
	\setlength{\tabcolsep}{1.5pt}
	\scalebox{0.85}{
		\begin{tabular}{ccccc}
			\toprule
			Reducing Setting & Params (M) & GMACs & Latency (ms) & Top-1 Acc. (\%)\\
			\midrule
			Baseline & 10.0 & 1.79 & 1.15 & 80.4 \\
			Number of Blocks & 8.4 & 1.70 & 1.07 & 79.7 \\
			FFN Width& 8.6 & 1.62 & 1.07 & 79.8 \\
			Attn. Head and FFN Width& 8.9 & 1.64 & 1.10 & 80.2 \\
			\bottomrule
		\end{tabular}
	}
	\vspace{-2mm}
\end{wraptable}
In contrast, we observe that an orchestrated reduction in both attention head and FFN expansion dimensions yields a milder accuracy decline (-0.2\%). These results demonstrate that a comprehensive reduction across different components offers better flexibility and performance.
\paragraph{Depthwise Convlution in FFN} \label{app:dw}
Recent works~\citep{cai2023efficientvit, qin2024mobilenetv4} attempt to insert a depthwise convolution (DW Conv) within the FFN to perform spatial mixing on the expanded features activations. We hypothesize that implementing more effective spatial mixing before the FFN diminishes its significance.
In our iFormer, depthwise convolution with a kernel size of 7 is employed for spatial modeling in the early layers, while a powerful SHMA is utilized in the later layers. This approach provides a significantly enhanced spatial mixing capacity than previous methods. 
\begin{wraptable}{r}{8.5cm}
	\caption{\textbf{Comparison of FFN with and without depthwise convolution.}}
	\label{tab:dw}
	\centering
	\vspace{-2mm}
	\small
	\setlength{\tabcolsep}{2.5pt}
	\scalebox{0.9}{
		\begin{tabular}{ccccc}
			\toprule
			DW Conv in FFN & Params (M) & GMACs & Latency (ms) & Top-1 Acc. (\%)\\
			\midrule
			with & 9.6 & 1.83 & 1.43 & 80.5 \\
			w/o. & 8.9 & 1.60 & 1.10 & 80.4\\
			\bottomrule
		\end{tabular}
	}
	\vspace{-2mm}
\end{wraptable}
As shown in Table~\ref{tab:dw}, enhancing all FFN with depthwise convolution, including those within the convolutional blocks, results in a +14\% increase in FLOPs and an additional latency cost of 0.33 ms. This increase is expected since the intermediate layers in the FFN possess an expanded feature dimension.
However, the Top-1 accuracy only exhibits a marginal improvement of +0.1\%.

\paragraph{Training for Longer Schedule}
Another commonly used advanced training is an extended schedule (450 \textit{vs.} 300). Here we provide additional experiments for both image classification and downstream tasks where we train iFormer with distillation for 450 epochs. To ensure a fair comparison with previous methods, we develop a larger model dubbed as iFormer-L2.
\begin{table}[h]
	\caption{\textbf{Training with distillation for 450 epochs on ImageNet-1K.}}
	\label{tab:longer cls}
	\centering
	\vspace{-2.5mm}
	\small
	\addtolength{\tabcolsep}{-2pt}
	\scalebox{0.8}{
		\begin{tabular}{cccccccccc}
			\toprule
			\multirow{1}{*}{Model} & Params (M) &  \multirow{1}{*}{\makecell{Latency (ms)}} & \multirow{1}{*}{Reso.}  & \multirow{1}{*}{Epochs} & \multirow{1}{*}{Top-1 (\%)} \\
			\midrule
			ConvNeXt-B~\citeyearpar{liu2022convnet} & 89.0 & 7.54 & 224 & 300 & 83.8 \\
			EfficientFormerV2-L~\citeyearpar{li2023rethinking} & 26.1 & 2.40  & 224 & 450 & 83.5 \\
			\rowcolor[gray]{0.92}
			\textbf{iFormer-L2} & \textbf{24.5}  & \textbf{2.30} &  224  & 450  & \textbf{83.9}\\
			\bottomrule
		\end{tabular}
	}
	\vspace{-2.5mm}
\end{table}
We report the image classification results on the ImageNet-1k dataset in Table~\ref{tab:longer cls}. It shows that training iFormer-L2 for 450 epochs yields improved performance, obtaining a Top-1 accuracy of 83.9\%, even surpassing the ConvNeXt-Base model.
\begin{table}[h]
	\centering
	\small
	\caption{
		\textbf{Object detection \& Semantic segmentation results using backbone pretrained for 450 epochs.}
	}
	\resizebox{0.95\linewidth}{!}{
		\begin{tabular}{c|c|c|c|ccc|ccc|c}
			\toprule
			\multirow{2}{*}{Backbone} &\multirow{2}{*}{\makecell{Param \\ (M)}} & \multirow{2}{*}{\makecell{Latency $\downarrow$ \\ (ms)}} & \multirow{2}{*}{Pretrain Epochs} &  \multicolumn{3}{c|}{Object Detection} & \multicolumn{3}{c|}{Instance Segmentation} & \multicolumn{1}{c}{Semantic}  \\ 
			\cmidrule{5-11} &  &  &  & AP$^\text{box}$ & AP$^\text{box}_{50}$   & AP$^\text{box}_{75}$   & AP$^\text{mask}$    & AP$^\text{mask}_{50}$   & AP$^\text{mask}_{75}$   & mIoU   \\
			\midrule
			\midrule
			ResNet50~\citeyearpar{he2016deep}  & 25.5 & 7.20 & 300 &38.0 & 58.6 & 41.4 & 34.4 & 55.1 & 36.7 & 36.7 \\
			PoolFormer-S24~\citeyearpar{yu2022metaformer}  &21.4    & 12.30  & 300 & 40.1   & 62.2    & 43.4    & 37.0   & 59.1    & 39.6  &  40.3 \\
			ConvNeXt-T~\citep{liu2022convnet} & 29.0 & 12.6 & 300 & 41.0 & 62.1 & 45.3 & 37.7 & 59.3 & 40.4 & 41.4 \\
			EfficientFormer-L3~\citeyearpar{li2022efficientformer}  & 31.3 & 8.40 & 300 & 41.4 & 63.9 & 44.7 & 38.1 & 61.0 & 40.4 & 43.5 \\
			RepViT-M1.5~\citeyearpar{wang2024repvit} & 14.0 & 5.00 & 300 & 41.6 & 63.2 & 45.3 & 38.6 & 60.5 & 41.5 & 43.6 \\
			PVTv2-B1~\citeyearpar{wang2022pvt} & 14.0 & 27.00 & 300 & 41.8 & 64.3 & 45.9 & 38.8 & 61.2 & 41.6 & 42.5 \\
			FastViT-SA24~\citeyearpar{vasu2023fastvit} & 20.6 & 8.97 & 300 & 42.0 & 63.5 & 45.8 & 38.0 & 60.5 & 40.5 & 41.0 \\
			EfficientMod-S~\citeyearpar{ma2024efficient} & 32.6 & 24.30 & 300 & 42.1 & 63.6 & 45.9 & 38.5 & 60.8 & 41.2 & 43.5 \\
			Swin-T~\citeyearpar{liu2021swin} & 28.3 & Failed & 300 & 42.2 & 64.4 & 46.2 & 39.1 & 61.6 & 42.0 & 41.5 \\
			\rowcolor[gray]{0.92}
			iFormer-L    &  \text{14.7} &   \text{6.60}  & 300  & \text{42.2}   & \text{64.2}    &  \text{46.0}   &  \text{39.1}   & \text{61.4}   & \text{41.9}     &   \text{44.5}   \\
			\midrule
			EfficientFormerV2-L~\citeyearpar{li2023rethinking}  & 26.1 & 12.5 & 450 & 44.7 & 66.3 & 48.8 & 40.4 & 63.5 & 43.2 & 45.2 \\
			\rowcolor[gray]{0.92}
			iFormer-L2    &  \text{24.5} &   \text{9.06}  & 450  & \text{44.6}   & \text{66.7}    &  \text{49.1}   &  \text{41.1}   & \text{64.0}   & \text{44.1}     &   \text{46.2}   \\
			\bottomrule
		\end{tabular}
	}
	\vspace{-2.5mm}
	\label{tab:longer coco}
\end{table}

Furthermore, we integrate iFormer-L2 into the Mask-RCNN and Semantic FPN framework for downstream tasks. As anticipated, the model with the more powerful iFormer-L2 backbone achieves SOTA performance, obtaining a significant enhancement over models pretrained for 300 epochs. It also outperforms its EfficientFormerV2-L counterpart by +0.7\% in AP$^\text{mask}$ and +1.0\% in mIoU, while being 1.4$\times$ faster. These experiments collectively show that advanced training strategies can be easily employed to improve the performance of iFormers.

\begin{figure*}[t]
	\centering
	\includegraphics[width=1\linewidth]{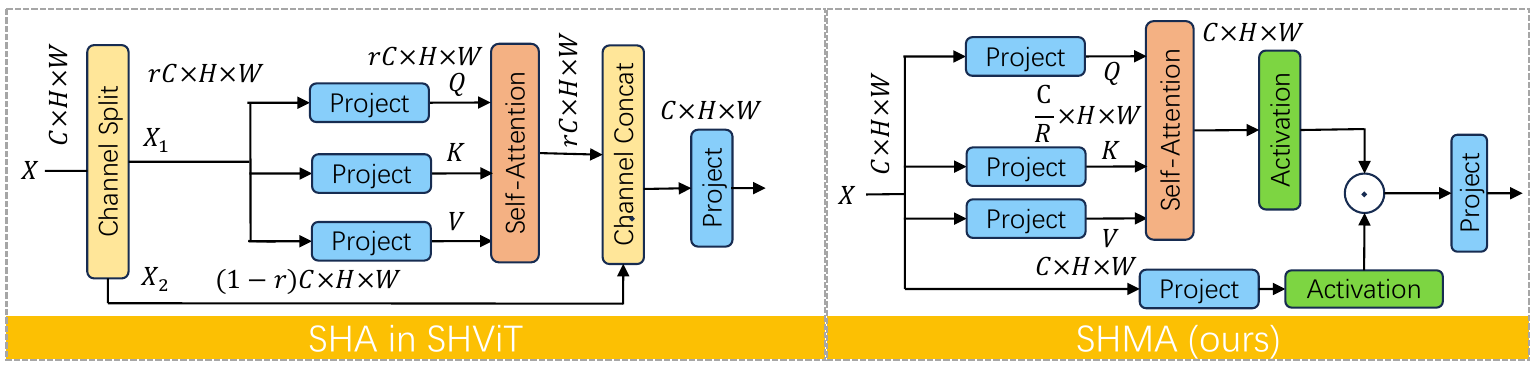}
	\vspace{-2mm}
	\caption{\textbf{Comparison of SHMA and SHA in SHViT.} In SHViT, $rC$ channels are utilized for spatial attention, where $r$ is set to $\frac{1}{4.67}$. SHMA projects the input into a higher dimension of $\frac{1}{2}$C (i.e., R=2) and avoids split and concatenation operations.}
	\label{fig:shvit}
	\vspace{-6mm}
\end{figure*}

\section{Relation to SHViT}\label{app:relation}
We clarify the difference between SHA in iFormer and its counterpart in SHViT~\citep{yun2024shvit} from the following two aspects: First, in terms of motivation, iFormer explores efficient attention mechanisms specifically tailored for the on-device environment, whereas SHViT is geared towards general-purpose GPUs, which may exhibit different hardware characteristics.
Second, in terms of methodology, as shown in Fig.~\ref{fig:shvit}, we utilize single-head attention with more channels ($R$ is set to 2.), while SHViT employs fewer than 1/4 of channels for attention. The reduced number of channels can result in a lower rank of the attention matrix, potentially degrading its expressiveness. Additionally, the split and concatenate operations in SHViT introduce extra runtime.

\begin{table}[h]
	\caption{\textbf{Process of converting SHA in iFormer towards SHViT.} Intermediate models are only measured by latency.}
	\label{tab:relation to shvit}
	\centering
	\small
	\vspace{-2mm}
	\scalebox{0.95}{
		\begin{tabular}{lcccc}
			\toprule
			Modification & Params(M) & GMACs & Latency (ms) & Top-1(\%) \\
			\toprule
			SHA Baseline without Modulation               &  9.9M  & 1758M &  1.12ms &  79.4 \\
			+ split                     &  9.9M  & 1758M &  1.18ms &     - \\
			+ attention on 1/4 channels &  8.3M  & 1547M &  1.02ms &     - \\
			+ concat                    &  8.7M  & 1579M &  1.11ms &   79.5 \\
			\bottomrule
		\end{tabular}
	}
	\vspace{-2mm}
\end{table}
We also conduct a more fair comparison with SHViT. We start from the SHA baseline referenced in Table~\ref{tab:single vs. multi head}, specifically denoted as 'SHA' in Figure~\ref{fig:roadmap}. The transition to SHViT involves the following steps: 1) splitting the input into two smaller tensors, $X_1$ and $X_2$, along the channel dimension; 2) applying single-head attention to the tensor $X_1$, which contains fewer than 1/4 of channels present in the original input tensor; and 3) concatenating the attention output with the residual input $X_2$. As summarized in Table~\ref{tab:relation to shvit}, split and concatenate operations introduce additional runtime. Furthermore, the performance of the SHA in the SHViT exhibits a decline compared to its counterpart in iFormer under similar latency conditions  (79.8 v.s. 79.5). This degraded performance may be attributed to the reduced number of channels in the attention mechanism.

\section{Architecture Details}\label{app:arch}

\begin{table*}[h]
	\small
	\centering
	\caption{\textbf{iFormer architecture configurations.} BN stands for Batch Normalization. SHMA stands for Singe-Head Modulation Attention. DW stands for Depthwise convolution. s and d means the stride and output dimension in convolution. hd denotes the head dimension in SHMA and the number of attention heads in all variants is 1. r means the expansion ratio in FFN.}
	\addtolength{\tabcolsep}{-2pt}
	\resizebox{.98\linewidth}{!}{
		\begin{tabular}{c|c|c|c|c|c}
			& \begin{tabular}[c]{@{}c@{}}Output Size \\ (Downs. Rate)\end{tabular} & iFormer-T  & iFormer-S & iFormer-M &  iFormer-L \\
			\hline
			\midrule
			\multirow{3}{*}{Stem} & \multirow{3}{*}{\begin{tabular}[c]{@{}c@{}}56$\times$56\\ (4$\times$)\end{tabular}} & $\begin{bmatrix}\text{Conv-BN-GELU 5$\times$5 s2 d16}\end{bmatrix}$ $\times$ 1 &
			$\begin{bmatrix}\text{Conv-BN-GELU 5$\times$5 s2 d16}\end{bmatrix}$ $\times$ 1 &
			$\begin{bmatrix}\text{Conv-BN-GELU 5$\times$5 s2 d24}\end{bmatrix}$ $\times$ 1 &
			$\begin{bmatrix}\text{Conv-BN-GELU 5$\times$5 s2 d24}\end{bmatrix}$ $\times$ 1 \\
			\cmidrule{3-6}
			& & $\begin{bmatrix}\text{Conv-BN-GELU 5$\times$5 s2 d64}\\\text{Conv-BN 1$\times$1 s1 d32} \end{bmatrix}$ $\times$ 1 &
			$\begin{bmatrix}\text{Conv-BN-GELU 5$\times$5 s2 d64}\\\text{Conv-BN 1$\times$1 s1 d32} \end{bmatrix}$ $\times$ 1 &
			$\begin{bmatrix}\text{Conv-BN-GELU 5$\times$5 s2 d96}\\\text{Conv-BN 1$\times$1 s1 d48} \end{bmatrix}$ $\times$ 1 &
			$\begin{bmatrix}\text{Conv-BN-GELU 5$\times$5 s2 d96}\\\text{Conv-BN 1$\times$1 s1 d48} \end{bmatrix}$ $\times$ 1  \\
			\midrule
			\multirow{1}{*}{Stage 1} & \multirow{1}{*}{\begin{tabular}[c]{@{}c@{}}56$\times$56\\ (4$\times$)\end{tabular}}
			& $\begin{bmatrix}\text{Conv-BN 7$\times$7 s1 d32}\\\text{Conv-BN-GELU 1$\times$1 s1 d96} \\\text{Conv-BN 1x1 s1 d32} \end{bmatrix}$ $\times$ 2  & 
			$\begin{bmatrix}\text{Conv-BN 7$\times$7 s1 d32}\\\text{Conv-BN-GELU 1$\times$1 s1 d128} \\\text{Conv-BN 1x1 s1 d32} \end{bmatrix}$ $\times$ 2 & 
			$\begin{bmatrix}\text{Conv-BN 7$\times$7 s1d48}\\\text{Conv-BN-GELU 1$\times$1 s1 d192} \\\text{Conv-BN 1x1 s1 d48} \end{bmatrix}$ $\times$ 2 & 
			$\begin{bmatrix}\text{Conv-BN 7$\times$7 s1 d48}\\\text{Conv-BN-GELU 1$\times$1 s1 d192} \\\text{Conv-BN 1x1 s1 d48} \end{bmatrix}$ $\times$ 2   \\
			\midrule
			\multirow{4}{*}{Stage 2} & \multirow{5}{*}{\begin{tabular}[c]{@{}c@{}}28$\times$28\\ (8$\times$)\end{tabular}} & $\begin{bmatrix}\text{Conv-BN 3$\times$3 s2 d64} \end{bmatrix}$ $\times$ 1  & 
			$\begin{bmatrix}\text{Conv-BN 3$\times$3 s2 d64} \end{bmatrix}$ $\times$ 1   & 
			$\begin{bmatrix}\text{Conv-BN 3$\times$3 s2 d96} \end{bmatrix}$ $\times$ 1  &
			$\begin{bmatrix}\text{Conv-BN 3$\times$3 s2 d96} \end{bmatrix}$ $\times$ 1   \\
			\cmidrule{3-6}
			& & $\begin{bmatrix}\text{Conv-BN 7$\times$7 s1 d64}\\\text{Conv-BN-GELU 1$\times$1 s1 d192} \\\text{Conv-BN 1x1 s1 d64} \end{bmatrix}$ $\times$ 2  & 
			$\begin{bmatrix}\text{Conv-BN 7$\times$7 s1 d64}\\\text{Conv-BN-GELU 1$\times$1 s1 d256} \\\text{Conv-BN 1x1 s1 d64} \end{bmatrix}$ $\times$ 2 & 
			$\begin{bmatrix}\text{Conv-BN 7$\times$7 s1 d96}\\\text{Conv-BN-GELU 1$\times$1 s1 d384} \\\text{Conv-BN 1x1 s1 d96} \end{bmatrix}$ $\times$ 2 & 
			$\begin{bmatrix}\text{Conv-BN 7$\times$7 s1 d96}\\\text{Conv-BN-GELU 1$\times$1 s1 d384} \\\text{Conv-BN 1x1 s1 d96} \end{bmatrix}$ $\times$ 2   \\
			\midrule
			\multirow{10}{*}{Stage 3} & \multirow{10}{*}{\begin{tabular}[c]{@{}c@{}}14$\times$14\\ (16$\times$)\end{tabular}} & $\begin{bmatrix}\text{Conv-BN 3$\times$3 s2 d128} \end{bmatrix}$ $\times$ 1  & 
			$\begin{bmatrix}\text{Conv-BN 3$\times$3 s2 d176} \end{bmatrix}$ $\times$ 1   & 
			$\begin{bmatrix}\text{Conv-BN 3$\times$3 s2 d192} \end{bmatrix}$ $\times$ 1  &
			$\begin{bmatrix}\text{Conv-BN 3$\times$3 s2 d256} \end{bmatrix}$ $\times$ 1   \\
			\cmidrule{3-6}
			& & $\begin{bmatrix}\text{Conv-BN 7$\times$7 s1 d128}\\\text{Conv-BN-GELU 1$\times$1 s1 d384} \\\text{Conv-BN 1$\times$1 s1 d128} \end{bmatrix}$ $\times$ 6  & 
			$\begin{bmatrix}\text{Conv-BN 7$\times$7 s1 d176}\\\text{Conv-BN-GELU 1$\times$1 s1 d704} \\\text{Conv-BN 1x1 s1 d176} \end{bmatrix}$ $\times$ 9 & 
			$\begin{bmatrix}\text{Conv-BN 7$\times$7 s1 d192}\\\text{Conv-BN-GELU 1$\times$1 s1 d768} \\\text{Conv-BN 1x1 s1 d192} \end{bmatrix}$ $\times$ 9 & 
			$\begin{bmatrix}\text{Conv-BN 7$\times$7 s1 d256}\\\text{Conv-BN-GELU 1$\times$1 s1 d1024} \\\text{Conv-BN 1x1 s1 d256} \end{bmatrix}$ $\times$ 8  \\
			\cmidrule{3-6}
			& & $\begin{bmatrix}\text{CPE 3$\times$3}\\\text{SHMA hd64}\\\text{FFN r2}\end{bmatrix}$ $\times$  3 & 
			$\begin{bmatrix}\text{CPE 3$\times$3}\\\text{SHMA hd88}\\\text{FFN r3}\end{bmatrix}$ $\times$  3 & 
			$\begin{bmatrix}\text{CPE 3$\times$3}\\\text{SHMA hd96}\\\text{FFN r3}\end{bmatrix}$ $\times$  4 & 
			$\begin{bmatrix}\text{CPE 3$\times$3}\\\text{SHMA hd128}\\\text{FFN r3}\end{bmatrix}$ $\times$  8  \\
			\cmidrule{3-6}
			& & $\begin{bmatrix}\text{Conv-BN 7$\times$7 s1 d128}\\\text{Conv-BN-GELU 1$\times$1 s1 d384} \\\text{Conv-BN 1x1 s1 d128} \end{bmatrix}$ $\times$ 1  & 
			$\begin{bmatrix}\text{Conv-BN 7$\times$7 s1 d176}\\\text{Conv-BN-GELU 1$\times$1 s1 d704} \\\text{Conv-BN 1$\times$1 s1 d176} \end{bmatrix}$ $\times$ 1 & 
			$\begin{bmatrix}\text{Conv-BN 7$\times$7 s1 d192}\\\text{Conv-BN-GELU 1$\times$1 s1 d768} \\\text{Conv-BN 1$\times$1 s1 d192} \end{bmatrix}$ $\times$ 1 & 
			$\begin{bmatrix}\text{Conv-BN 7$\times$7 s1 d256}\\\text{Conv-BN-GELU 1$\times$1 s1 d1024} \\\text{Conv-BN 1$\times$1 s1 d256} \end{bmatrix}$ $\times$ 1  \\
			\midrule
			\multirow{4}{*}{Stage 4} & \multirow{4}{*}{\begin{tabular}[c]{@{}c@{}}7$\times$7\\ (32$\times$)\end{tabular}} 
			& $\begin{bmatrix}\text{Conv-BN 3$\times$3 s2 d256} \end{bmatrix}$ $\times$ 1  & 
			$\begin{bmatrix}\text{Conv-BN 3$\times$3 s2 d320} \end{bmatrix}$ $\times$ 1   & 
			$\begin{bmatrix}\text{Conv-BN 3$\times$3 s2 d384} \end{bmatrix}$ $\times$ 1  &
			$\begin{bmatrix}\text{Conv-BN 3$\times$3 s2 d384} \end{bmatrix}$ $\times$ 1   \\
			\cmidrule{3-6}
			& & $\begin{bmatrix}\text{CPE 3$\times$3}\\\text{SHMA hd64}\\\text{FFN r2}\end{bmatrix}$ $\times$  2 & 
			$\begin{bmatrix}\text{CPE 3$\times$3}\\\text{SHMA hd80}\\\text{FFN r3}\end{bmatrix}$ $\times$  2 & 
			$\begin{bmatrix}\text{CPE 3$\times$3}\\\text{SHMA hd96}\\\text{FFN r3}\end{bmatrix}$ $\times$  2 & 
			$\begin{bmatrix}\text{CPE 3$\times$3}\\\text{SHMA hd96}\\\text{FFN r3}\end{bmatrix}$ $\times$  2  \\
			\midrule
			\multicolumn{2}{c|}{Params (M)} & 2.9 & 6.5 & 8.9 & 14.7 \\
			\midrule			
			\multicolumn{2}{c|}{GMacs} & 0.53 & 1.09 & 1.64 & 2.63 \\
			\bottomrule
		\end{tabular}
	}
	\normalsize
	\vspace{-2.5mm}
	\label{tab:arch spec}
\end{table*}
In Table~\ref{tab:arch spec}, we show the different architecture configurations of the iFormer model variants. 

\section{iFormer for Higher Resolution}
\begin{table*}[h]
	\small
	\centering
	\caption{\textbf{Comparison of different attention designs in iFormer-M.} For the sake of simplicity, we exclude other blocks that are not related to attention. ws is the window size for window attention.}
	\addtolength{\tabcolsep}{-2pt}
	\resizebox{.9\linewidth}{!}{
		\begin{tabular}{c|c|c|c|c|}
			& Attention & SHMA & Hybrid SHMA & Chunk Hybrid SHMA\\
			\midrule
			\multirow{9}{*}{Stage 3} & \multirow{9}{*}{\begin{tabular}[c]{@{}c@{}}14$\times$14\\ (16$\times$)\end{tabular}} &
			& $\begin{bmatrix}\text{CPE 3$\times$3}\\\text{Window Partitioning, ws16}\\\text{Window SHMA hd96, ws16}\\\text{FFN r3}\end{bmatrix}$ $\times$  1  &
			$\begin{bmatrix}\text{CPE 3$\times$3}\\\text{Chunk Window Partitioning, ws16}\\\text{Window SHMA hd96, ws16}\\\text{FFN r3}\end{bmatrix}$ $\times$  1 \\
			\cmidrule{3-4}
			& & $\begin{bmatrix}\text{CPE 3$\times$3}\\\text{SHMA hd96}\\\text{FFN r3}\end{bmatrix}$ $\times$ 4  &
			$\begin{bmatrix}\text{CPE 3$\times$3}\\\text{Window SHMA hd96, ws16}\\\text{FFN r3}\end{bmatrix}$ $\times$  2  &
			$\begin{bmatrix}\text{CPE 3$\times$3}\\\text{Window SHMA hd96, ws16}\\\text{FFN r3}\end{bmatrix}$ $\times$  2 \\
			\cmidrule{3-4}
			& & & 			
			$\begin{bmatrix}\text{CPE 3$\times$3}\\\text{Window Reversing, ws16}\\\text{SHMA hd96}\\\text{FFN r3}\end{bmatrix}$ $\times$  1  &
			$\begin{bmatrix}\text{CPE 3$\times$3}\\\text{Chunk Window Reversing, ws16}\\\text{SHMA hd96}\\\text{FFN r3}\end{bmatrix}$ $\times$  1  \\
			\midrule
			\multirow{6}{*}{Stage 4} & \multirow{6}{*}{\begin{tabular}[c]{@{}c@{}}7$\times$7\\ (32$\times$)\end{tabular}} 
			& 
			& $\begin{bmatrix}\text{CPE 3$\times$3}\\\text{Window Partitioning, ws16}\\\text{Window SHMA hd96}\\\text{FFN r3}\end{bmatrix}$ $\times$  1  & 
			$\begin{bmatrix}\text{CPE 3$\times$3}\\\text{Chunk Window Partitioning, ws16}\\\text{Window SHMA hd96}\\\text{FFN r3}\end{bmatrix}$ $\times$  1 \\
			\cmidrule{3-4}
			& &  $\begin{bmatrix}\text{CPE 3$\times$3}\\\text{SHMA hd64}\\\text{FFN r2}\end{bmatrix}$ $\times$  2 & 
			$\begin{bmatrix}\text{CPE 3$\times$3}\\\text{Window Reversing, ws16}\\\text{SHMA hd64}\\\text{FFN r3}\end{bmatrix}$ $\times$  1  &
			$\begin{bmatrix}\text{CPE 3$\times$3}\\\text{Chunk Window Reversing, ws16}\\\text{SHMA hd64}\\\text{FFN r3}\end{bmatrix}$ $\times$  1 \\
			\bottomrule
		\end{tabular}
	}
	\normalsize
	\vspace{-2.5mm}
	\label{tab:attention comparison}
\end{table*}

Self-attention exhibits quadratic complexity with respect to the number of tokens, \textit{i.e.}, the resolution of the input image. This issue is exacerbated in dense prediction tasks, which usually require high-resolution input such as 512$\times$512 in semantic segmentation and generate a large amount of 1024 image tokens even in the third stage. Consequently, this will cause huge memory and computation costs in mobile devices. 

\begin{wraptable}{r}{8cm}
	\caption{\textbf{Latency comparison of different attention mechanisms.}}
	\label{tab:higher res}
	\centering
	\vspace{-2mm}
	\small
	\scalebox{0.95}{
		\begin{tabular}{ccc}
			\toprule
			Attention & Resolution & Latency (ms) \\
			\toprule
			SHMA & 224 &  1.10 \\
			SHMA & 512 & Failed \\
			Hybrid SHMA & 512 & 11.46 \\
			CC Hybrid SHMA & 512 & 4.0 \\
			\bottomrule
		\end{tabular}
	}
	\vspace{-2mm}
\end{wraptable}
To mitigate these issues, we resort to window attention as proposed in Swin~\citep{liu2021swin}. However, default window attention only performs local self-attention within windows, thus lacking interactions between tokens from different windows which will impair modeling capacity. 
Swin introduces shifted window attention to alleviate this limitation. Unfortunately, the shifting operation inevitably incurs additional memory costs. In contrast to Swin, we implement a hybrid attention design. Specifically, we compute window attention within windows, except for the last attention block in each stage. This approach enables iFormer to capture more global features essential for dense prediction tasks. At the same time, since window partitioning and reversing also incur memory access costs, we minimize the usage of them to once per stage. We replace the standard SHMA in iFormer with a hybrid window SHMA, as shown in Table~\ref{tab:attention comparison}. 

From the latency comparison in Table~\ref{tab:higher res}, we see that simply applying SHMA will encounter a memory bottleneck on mobile devices. Instead, our hybrid SHMA can significantly reduce memory access costs, achieving a mobile latency of 11.46 ms. 

However, hybrid SHMA still lags much behind the recent FastViT-SA12, which has a latency of 5.27 ms. We identify the speed bottleneck as stemming from the window partitioning and reversing operations, even though we only implement them once in each stage. As the feature map size increases, the reshaping involved in these operations demands considerable memory, thereby slowing inference in resource-constrained mobile devices. 

To address this issue, we propose a method called ``Channel Chunking" (CC). Formally, given a 2D input feature map $\mathbf{x}\in \mathbb{R}^{C\times H\times W}$, the standard window partitioning divides the feature map into $\frac{H}{P}\times \frac{W}{P}$ non-overlapped regions, each corresponding to a window that contains $P\times P$ feature vectors. This step is accomplished by reshaping x as $\mathbf{x^P}\in  \mathbb{R}^{\frac{HW}{P^2}\times C\times P\times P}$. Then we apply SHMA within each window. 

To reduce the memory requirements associated with reshaping, we propose to split the feature map $\mathbf{x}$ along the channel dimension into a series of smaller chunks as follows:
\begin{align}
	\begin{split}
		\mathbf{x_1^S}, ..., \mathbf{x_n^S} = \text{Chunking}(\mathbf{x}),
	\end{split}
\end{align}
where K is the chunk size, $n=\frac{C}{K}$ is the number of chunks. We set n=16 for the input image of 512$\times$512 in our object detection and semantic segmentation experiments.
Then we apply window partitioning sequentially to these smaller chunks and concatenate them. This process can be mathematically expressed as follows:
\begin{align}
	\begin{split}
		& \mathbf{x^P} = \text{Concat}(\mathbf{x^P_i} , ..., \mathbf{x^P_n}), \\
		& \text{where} \quad \mathbf{x^P_i} = \text{WindowPartitioning}(\mathbf{x^S_i}),
	\end{split}
\end{align}
These smaller chunks can be processed rapidly. As shown in Table~\ref{tab:higher res}, the chunking strategy allows the model to achieve 2.9$\times$ speed up in inference speed.
Correspondingly, the window reversing operation is performed by reshaping multiple windows $\mathbf{x^P} \in  \mathbb{R}^{\frac{HW}{P^2}\times C\times P\times P}$ into a 2D feature map $\mathbf{x}\in \mathbb{R}^{C\times H\times W}$. These results demonstrate that our proposed Channel Chunking Hybrid SHMA significantly enhances the iFormer’s ability to process high-resolution images efficiently.

\paragraph{Computation Complexity}
Given an input $\ \mathbf{x}\in \mathbb{R}^{C\times H\times W}$ and a window size of P $\times$ P, as detailed in Section E, the computational complexity of iFormer is as follows:
\begin{align}
		\begin{split}
		\Omega (\text{SHMA}) = & 4HWC^2 \text{(QKV and output projection)} + \\
													    & HWC \text{(element-wise product of modulation)} +  \\
														& 2P^2HWC \text{(self-attention)}, \\
		\end{split}
\end{align}
\begin{align}
	\begin{split}
		\Omega (\text{FFN}) = 8HWC^2.
		\end{split}
\end{align}
In image classification, we do not utilize window attention since the feature size is 14$\times$ 14 in stage 3 (it equals to the window attention when P=14). In downstream tasks, we adopt a window size of P=16.

\section{Comprehensive Comparison} \label{app:comprehensive comparison}
\begin{table}[t]
	\caption{{\textbf{Comprehensive comparison between iFormer and the previously proposed models on ImageNet-1K. } Failed indicated that the model runs too long to report latency by the Core ML, often caused by excessive memory access.}}
	\label{tab:app image classification}
	\centering
	\vspace{-2mm}
	\small
	\scalebox{0.9}{
		\begin{tabular}{cccccccccc}
			\toprule
			\multirow{2}{*}{Model}   & \multirow{2}{*}{Params (M)} & \multirow{2}{*}{GMACs} &  \multirow{2}{*}{\makecell{Latency $\downarrow$ \\ (ms)}}  & \multirow{2}{*}{Reso.}   & \multirow{2}{*}{Epochs} & \multirow{2}{*}{Top-1 (\%)}  \\
			\\
			\midrule
			MobileNetV2 1.0x~\citeyearpar{sandler2018mobilenetv2} & 3.4 & 0.30 & 0.73 & 224 & 500 & 72.0 \\
			SHViT-S1~\citeyearpar{yun2024shvit} & 6.3 & 0.24 & 0.74 & 224 & 300 & 72.8 \\
			MobileNetV3-Large 0.75x~\citeyearpar{howard2019searching} & 4.0 & 0.16 & 0.67 & 224 & 600 & 73.3 \\
			MNV4-Conv-S~\citeyearpar{qin2024mobilenetv4} & 3.8 & 0.20 & 0.60 & 224 & 500 & 73.8 \\
			\rowcolor[gray]{0.92}
			\textbf{iFormer-T} & 2.9& 0.53 & \textbf{0.60} &  224  & 300  & \textbf{74.1}\\
			\midrule
			ShuffleNetV2 1.0×~\citeyearpar{ma2018shufflenet} & 2.3 & 0.15 & 0.74 & 224 & 300  & 69.4 \\ 
			MobileNetV2 1.4x~\citeyearpar{sandler2018mobilenetv2} & 6.9 & 0.59 & 1.02 & 224 & 500 & 74.7 \\
			MobileNetV3-Large 1.0x~\citeyearpar{howard2019searching} & 5.4 & 0.22 & 0.76 & 224 & 600 & 75.2 \\
			SwiftFormer-XS~\citeyearpar{shaker2023swiftformer}&3.5 & 0.60 & 0.95 & 224 & 300 & 75.7 \\
			SBCFormer-XS~\citeyearpar{lu2024sbcformer} & 5.6 & 0.70 & 0.79 & 224 & 300 & 75.8  \\
			${\text{GhostNetV3 1.0x}^{\dag}}$~\citeyearpar{liu2024ghostnetv3} & 6.1 & 0.17 & 0.99 &224 & 600 & 77.1 \\
			EfficientNet-B0~\citeyearpar{tan2019efficientnet} & 5.3 & 0.39 & 0.89 & 224 & 350 & 77.1 \\
			MobileOne-S2~\citeyearpar{vasu2023mobileone} & 7.8 & 1.30 & 0.92 & 224 & 300 & 77.4 \\
			LowFormer-B0~\citeyearpar{nottebaum2024lowformer} & 14.1 & 0.94 & 1.45 & 224 & 300 & 78.4  \\
			CAS-ViT-XS~\citeyearpar{zhang2024cas} & 3.2 & 0.56 & 0.85 & 224 & 300 & 77.5 \\
			EMO-5M~\citeyearpar{zhang2023rethinking} &5.1 & 0.90 & Failed & 224 & 300 & 78.4 \\
			RepViT-M1.0~\citeyearpar{wang2024repvit}  & 6.8 & 1.10 & 0.85 & 224 & 300 & 78.6 \\
			\rowcolor[gray]{0.92}
			\textbf{iFormer-S} & 6.5 & 1.09 & \textbf{0.85} &  224  & 300  & \textbf{78.8}\\
			\midrule
			ShuffleNetV2 1.5×~\citeyearpar{ma2018shufflenet} & 3.5 & 0.30 & 1.16 & 224 & 300 & 72.6 \\
			EdgeViT-XXS~\citeyearpar{pan2022edgevits} & 4.1 & 0.60 & 1.41 & 224 & 300 & 74.4 \\
			SHViT-S2~\citeyearpar{yun2024shvit} & 11.4 & 0.37 & 1.10 & 224 & 300 & 75.2 \\
			EfficientMod-xxs~\citeyearpar{ma2024efficient} & 4.7 & 0.60 & 1.29 & 224 & 300 & 76.0 \\
			SBCFormer-S~\citeyearpar{lu2024sbcformer} & 8.5 & 0.90 & 1.02 & 224 & 300 & 77.7 \\
			MobileOne-S3~\citeyearpar{vasu2023mobileone} & 10.1 & 1.90 & 1.16 & 224 & 300 & 78.1  \\
			SwiftFormer-S~\citeyearpar{shaker2023swiftformer}&6.1 & 1.00 & 1.12 & 224 & 300 & 78.5 \\
			${\text{GhostNetV3 1.3x}^{\dag}}$~\citeyearpar{liu2024ghostnetv3}  & 8.9 & 0.27 & 1.24 & 224 & 600 & 79.1 \\
			EfficientNet-B1~\citeyearpar{tan2019efficientnet} & 7.8 & 0.70 & 1.29 & 240 & 350 & 79.1 \\
			FastViT-T12~\citeyearpar{vasu2023fastvit} & 6.8 & 1.40 & 1.12 & 256 & 300 & 79.1 \\
			RepViT-M1.1~\citeyearpar{wang2024repvit}  & 8.2 & 1.30 & 1.04 & 224 & 300 & 79.4 \\
			RepNeXt-M3~\citeyearpar{zhao2024repnext} & 7.8 & 1.30 & 1.04 & 224 & 300 &  79.4\\
			FastViT-S12~\citeyearpar{vasu2023fastvit} & 8.8 & 1.80 & 1.26 & 256 & 300 & 79.8 \\
			MNV4-Conv-M~\citeyearpar{qin2024mobilenetv4} & 9.2 &1.00 & 1.08 & 256 & 500 & 79.9 \\
			\rowcolor[gray]{0.92}
			\textbf{iFormer-M} & 8.9 & 1.64 & \textbf{1.10} &  224  & 300  & \textbf{80.4} \\
			\midrule
			MobileViT-XXS~\citeyearpar{mehta2021mobilevit}&1.3 & 0.40 & 2.12 & 256 & 300 & 69.0 \\
			MobileViTV2-0.5~\citeyearpar{mehta2022separable} & 1.4 & 0.50 & 9.47 & 256 & 300 & 70.2 \\
			ShuffleNet v2 2.0×~\citeyearpar{ma2018shufflenet} & 7.4 & 0.59 & 1.94 & 224 & 300  & 74.9 \\ 
			EdgeViT-XS~\citeyearpar{pan2022edgevits} & 6.7 & 1.10 & 1.79 & 224 & 300 & 77.5 \\
			Mobile-Former-294M~\citeyearpar{chen2022mobile} &11.4 & 0.29 & 2.66 & 224 & 450 & 77.9 \\
			MobileViTV2-1.0~\citeyearpar{mehta2022separable} & 4.9 & 1.80 & Failed & 256 & 300 & 78.1 \\
			EfficientMod-xs~\citeyearpar{ma2024efficient} & 6.6 & 0.80 & 2.13 & 224 & 300 & 78.3 \\
			MobileViT-S~\citeyearpar{mehta2021mobilevit}& 5.6 & 2.00 & 3.55 & 256 & 300 & 78.4 \\
			CMT-Ti~\citeyearpar{guo2022cmt} & 11.3 & 687 & Failed & 160 & 300 & 79.2\\
			Mobile-Former-508M~\citeyearpar{chen2022mobile} &14 & 0.51 & 3.33 & 224 & 450 & 79.3 \\
			SHViT-S4~\citeyearpar{yun2024shvit} & 16.5 & 0.99 & 1.48 & 224 & 300 & 79.4 \\
			EfficientViT-B1-r224~\citeyearpar{cai2023efficientvit} & 9.1 & 0.52 & 2.38 & 224 & 350 & 79.4 \\
			MobileOne-S4~\citeyearpar{vasu2023mobileone} & 14.8 & 2.98 & 1.74 & 224 & 300 & 79.4  \\
			LowFormer-B1~\citeyearpar{nottebaum2024lowformer} & 17.9 & 1.41 & 1.90 & 224 & 300 & 79.9  \\
			SBCFormer-B~\citeyearpar{lu2024sbcformer} & 13.8 & 1.60 & 1.44 & 224 & 300 & 80.0 \\
			EfficientNet-B2~\citeyearpar{tan2019efficientnet} & 9.2 & 1.00 & 1.69 & 260 & 350 & 80.1 \\
			CAS-ViT-S~\citeyearpar{zhang2024cas} & 5.8 & 0.93 & 1.82 & 224 & 300 & 80.2 \\
			${\text{GhostNetV3 1.6x}^{\dag}}$~\citeyearpar{liu2024ghostnetv3} & 12.3 & 0.40 & 1.49 & 224 & 600 & 80.4 \\
			EfficientViT-B1-r288~\citeyearpar{cai2023efficientvit} & 9.1 & 0.86 & 3.87 & 288 & 450 & 80.4 \\
			FastViT-SA12~\citeyearpar{vasu2023fastvit} & 10.9 & 1.90 & 1.50 & 256 & 300 & 80.6 \\
			MNV4-Hybrid-M~\citeyearpar{qin2024mobilenetv4} & 10.5  & 1.20 & 1.75 & 256 & 500 & 80.7 \\
			SwiftFormer-L1~\citeyearpar{shaker2023swiftformer}& 12.1 & 1.60 & 1.60 & 224 & 300 & 80.9 \\
			EfficientMod-s~\citeyearpar{ma2024efficient} & 12.9 & 1.40 & 2.57 & 224 &300 & 81.0 \\
			SBCFormer-L~\citeyearpar{lu2024sbcformer} & 18.5 & 2.70 & 1.89 & 224 & 300 & 81.1 \\
			RepViT-M1.5~\citeyearpar{wang2024repvit}  & 14.0 & 2.30& 1.64 & 224 & 300 & 81.2 \\
			LowFormer-B1.5~\citeyearpar{nottebaum2024lowformer} & 33.9 & 2.57 & 3.02 & 224 & 300 & 81.2 \\
			RepNeXt-M4~\citeyearpar{zhao2024repnext} & 13.3 & 2.30 & 1.47 & 224 & 300 & 81.2 \\
			CAS-ViT-M~\citeyearpar{zhang2024cas} & 12.4 & 1.89 & 2.46 & 224 & 300 & 81.4 \\
			\rowcolor[gray]{0.92}
			\textbf{iFormer-L} & 14.7 & 2.63 & \textbf{1.60} &  224  & 300  & \textbf{81.7}\\
			\bottomrule
		\end{tabular}
	}
	\vspace{-2.5mm}
\end{table}
In Table~\ref{tab:app image classification}, we provide a more comprehensive comparison between iFormer and other lightweight models on ImageNet-1k classification.
\end{document}